%% file: main.tex
\documentclass[11pt]{style/prism-princeton}

\input{preamble/notation}

\author[1*]{Arhan Jain}
\author[2*, \ensuremath{\alpha}]{Mingtong Zhang}
\author[1]{Kanav Arora}
\author[3]{William Chen}
\author[4]{Marcel Torne}
\author[5]{Muhammad Zubair Irshad}
\author[5]{Sergey Zakharov}
\author[6]{Yue Wang}
\author[3, 8]{Sergey Levine}
\author[4, 8]{Chelsea Finn}
\author[7]{Wei-Chiu Ma}
\author[2]{Dhruv Shah}
\author[1,5\ddagger]{Abhishek Gupta}
\author[3,4,8\ddagger]{Karl Pertsch}

\affiliation[1]{University of Washington}
\affiliation[2]{Princeton University}
\affiliation[3]{University of California, Berkeley}
\affiliation[4]{Stanford University}
\affiliation[5]{Toyota Research Institute}
\affiliation[6]{University of Southern California}

\affiliation[7]{Cornell University}
\affiliation[8]{Physical Intelligence}

\contribution[*]{Equal contribution.}
\contribution[\ddagger]{Equal advising.}
\contribution[\ensuremath{\alpha}]{Work partially done at TRI.}

\usepackage{wrapfig}
\usepackage[most]{tcolorbox}
\usepackage{xcolor}

\definecolor{polaris-bg-elevated}{HTML}{f5f5f5}
\definecolor{polaris-border-subtle}{HTML}{e5e5e5}

\def \Acronym {\textsc{PolaRiS}}
\begin{document}

\title{\Acronym{}: Scalable Real-to-Sim Evaluations for Generalist Robot Policies}

\input{sections/00_abstract}

\website{
https://polaris-evals.github.io  %
}
{
polaris-evals.github.io   %
}

\maketitle

\input{sections/01_introduction}
\input{sections/02_related}
\input{sections/03_prelims}
\input{sections/04_methods}

\input{sections/05_experiments}
\input{sections/06_discussion}
\input{sections/07_acks}

\clearpage

\bibliographystyle{unsrt} %
\bibliography{references}

\clearpage

\beginappendix{
\input{sections/X_supp}
}

\end{document}

%% file: preamble/notation.tex
\makeatletter
\newcommand{\longdash}[1][2em]{%
  \makebox[#1]{$\m@th\smash-\mkern-7mu\cleaders\hbox{$\mkern-2mu\smash-\mkern-2mu$}\hfill\mkern-7mu\smash-$}}
\makeatother
\newcommand{\omitskip}{\kern-\arraycolsep}

%% file: sections/00_abstract.tex
\abstract{
A significant challenge for robot learning research is our ability to accurately measure and compare the performance of robot policies. Benchmarking in robotics is historically challenging due to the stochasticity, reproducibility, and time-consuming nature of real-world rollouts. This challenge is exacerbated for recent generalist policies, which has to be evaluated across a wide variety of scenes and tasks. Evaluation in simulation offers a scalable complement to real world evaluations, but the visual and physical domain gap between existing simulation benchmarks and the real world has made them an unreliable signal for policy improvement. Furthermore, building realistic and diverse simulated environments has traditionally required significant human effort and expertise. To bridge the gap, we introduce \textbf{Policy Evaluation and Environment Reconstruction in Simulation (\Acronym{})}, a scalable real-to-sim framework for high-fidelity simulated robot evaluation. \Acronym{} utilizes neural reconstruction methods to turn short video scans of real-world scenes into interactive simulation environments. Additionally, we develop a simple simulation data co-training recipe that bridges remaining real-to-sim gaps and enables zero-shot evaluation in \textit{unseen} simulation environments. Through extensive paired evaluations between simulation and the real world, we demonstrate that \Acronym{} evaluations provide a much stronger correlation to real world generalist policy performance than existing simulated benchmarks. Its simplicity also enables rapid creation of diverse simulated environments. 
As such, this work takes a step towards distributed and democratized evaluation for the next generation of robotic foundation models.
}

%% file: sections/01_introduction.tex
\begin{figure*}
    \centering
    \includegraphics[width=\textwidth]{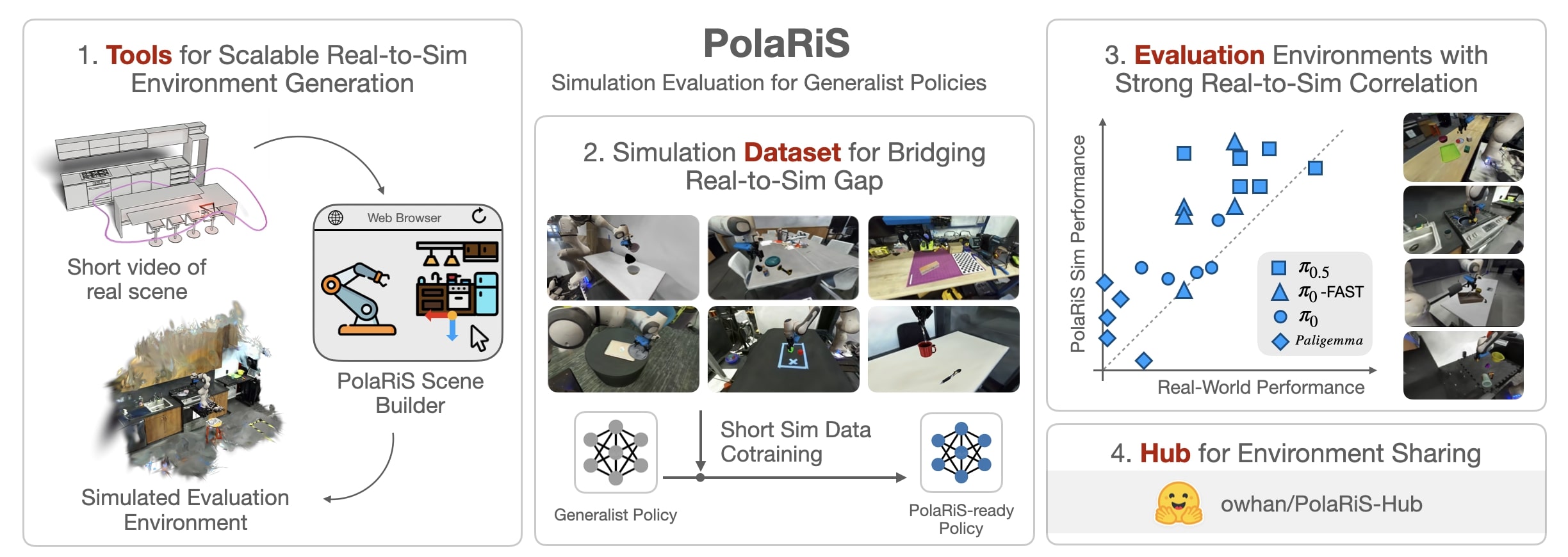}
    \caption{\footnotesize{\textbf{Overview of \Acronym{}.} \Acronym{} is a real-to-sim approach that turns short videos of real-world scenes into high-fidelity simulated environments for scalable robot policy evaluation. We demonstrate that co-finetuning with simulation data for a small number of iterations enables generalist policies to be evaluated accurately in \emph{unseen} \Acronym{} environments \emph{out of the box}. As part of this work, we provide: (1)~\textbf{tools} for quickly creating new \Acronym{} environments, (2)~an open-source simulation \textbf{dataset} for finetuning generalist policies for \Acronym{} evaluation, (3)~off-the-shelf \textbf{evaluation environments} with strong real-to-sim correlation, and (4)~a \textbf{hub} for sharing new \Acronym{} environments with the community.}}
    \label{fig:teaser}
\end{figure*}

\section{Introduction}
\label{sec:intro}
Recently, \emph{generalist robotic policies}~\cite{blackpi0, pi05, o2024open,  team2025gemini, geminirobotics, lbmtri2025, octo_2023, kim24openvla}, trained on diverse real-robot datasets, have shown an impressive degree of generalization to novel scenarios. However, with the advent of generalist policies comes a critical demand for large-scale empirical validation, since absolute guarantees are challenging to provide. To enable generalization, evaluations must be conducted across a large variety of tasks, environments and rollouts. While performing evaluations directly on real-world robots is the gold standard, this quickly becomes tedious, hard to control systematically, and impractical for individual groups to perform due to cost and labor constraints. Therefore, a more scalable paradigm for generalist robot policy evaluation becomes an emerging demand. 

Simulation provides an appealing alternative---cheap, safe, fast, and unburdened by physical hardware, making large-scale policy evaluation far more feasible. However, despite considerable efforts~\cite{li24simpler, katara24gensim, liu23libero, nasiriany24robocasa, james20rlbench, mees22calvin, yu2020meta, li2024behavior},  these platforms have not been well suited for evaluating the aforementioned \emph{generalist} policies. To support a wide range of scenarios required to test generality, existing robot-simulation benchmarks typically sacrifice realism---including dynamics, visuals, and scene geometry---and thus fail to capture the visual complexity of the physical world ~\cite{li24simpler, yu19metaworld, mees22calvin, liu23libero, nasiriany24robocasa}. Our experiments show that there remains a poor correlation between simulation and real world policy performance of state-of-the-art generalist policies. Moreover, as generalist policies get deployed across diverse environments, roboticists are often interested not only in \emph{average} model performance \emph{across} environments, but also in understanding a policy's performance within a \emph{specific target environment}, which necessitates a tailored simulation of that scene.  Constructing such an environment typically requires considerable manual effort. Effective simulated evaluation of generalist policies requires a new set of tools that allow us to rapidly construct high-fidelity targeted simulation environments, reflecting the diversity of real-world environments.

To this end, we introduce \Acronym{}, a scalable real-to-sim framework for evaluating generalist robot policies. \Acronym{} leverages neural reconstruction such as Gaussian splatting~\cite{2dgs, 3dgs} and 3D object generation~\cite{xiang2025structured} to construct simulated evaluation environments that closely match their real-world counterparts with high visual and geometric fidelity (\Cref{fig:teaser}). This enables advanced generalist policies to perform various tasks in \Acronym{} simulations out of the box. Additionally, our experiments show that lightweight finetuning of these policies with a small dataset of demonstrations in simulation can significantly boost real-to-sim correlation. Importantly, unlike existing simulation benchmarks, this simulation data is used only to align visual representations instead of teaching new tasks, and therefore does not have to be collected in the target environment or for the target task. As a result, once finetuned, policies can be evaluated \emph{zero-shot} in \emph{unseen} \Acronym{} environments, facilitating rapid development of simulated evaluations that are customized for individual deployments yet shared broadly to assess generalist competence.

In our experiments across diverse real-world environments using the DROID platform~\cite{khazatsky24droid}, \Acronym{} evaluations match the real-world performance of state-of-the-art generalist policies out of the box with an average Pearson correlation of $r=0.9$. \Acronym{} evaluations also strongly correlate ($r=0.98$) with policy scores in RoboArena~\citep{atreya2025roboarena}, the most comprehensive real-world benchmark for generalist robot policies to date. 
As such, our experiments suggest that \Acronym{} takes a significant step towards rapid, scalable, simulated evaluation of generalist policies. We open-source all tools required to create, use, and share \Acronym{} evaluation environments, to foster the community-driven development of a shared suite of realistic simulation evaluation environments.

The contributions of this paper are as follows: (1)~we introduce \Acronym{}, a scalable framework for creating high-fidelity simulated evaluation environments through neural scene reconstruction, (2)~we develop a simple recipe for simulation data co-training that bridges the real-to-sim domain gap, ensuring that real-world generalist policies are accurately evaluatable in unseen simulation scenes out of the box, (3)~we perform empirical evaluations across diverse real-world scenes demonstrating strong performance correlation for widely used generalist policies between \Acronym{} simulation and the real world, (4)~we conduct detailed ablations demonstrating key design decisions for faithful simulated evaluation.

%% file: sections/02_related.tex
\section{Related Work}
\label{sec:related}

\textbf{Robot Policy Evaluation:}
Accurately evaluating the performance of robot policies is important for guiding research, yet faces challenges in practice: from reproducing initial environment conditions to robot wear and tear, and the tedium of supervising many hours of evaluations. Nevertheless, numerous real-robot benchmarks have been proposed over the years ~\cite{calli15ycb, ahn19robel, zhou25autoeval, atreya2025roboarena}. With the rise of increasingly general policies that can perform many tasks across many environments ~\cite{pi05, blackpi0, geminirobotics, lbmtri2025, nvidia2025gr00tn1openfoundation}, the challenges for performing comprehensive real-world evaluations are exacerbated. \emph{Simulated} evaluations pose a promising alternative: initial conditions in simulation are perfectly repeatable, simulated robots do not wear out, and evaluations can be rapidly scaled through parallelization. As a result, many simulated robot benchmarks have been proposed~\cite{li24simpler, liu23libero, nasiriany24robocasa, james20rlbench, yu19metaworld}. These benchmarks are invaluable for algorithmic development. However, due to large visual and physical domain gaps between these simulated environments and the real world, they struggle to evaluate the performance of \emph{generalist} policies, trained on real-world data. We instead design simulated environments that are capable of evaluating \emph{real}-robot, generalist policies. They share the aforementioned benefits of simulated evaluations with respect to usability, reproducibility, and scalability, but are also reflective of the real-world performance of the evaluated generalist policies. 
More recently, advances in world models have enabled the use of action-conditioned video models as an alternative tool for robot policy evaluation~\cite{quevedo2025evaluating, tseng2025scalable, guo2025ctrl, veorobotics2025}. Like ``classic'' simulation, video model evaluations can be run at scale. Yet, our experiments demonstrate that today's (open) models struggle to reliably model physical interactions, and thus struggle to estimate policy performance. In contrast, our method leverages an accurate physics simulator together with high-fidelity neural rendering, making policy evaluation more faithful to real-world dynamics, reliable, efficient, and cost-effective.

\noindent\textbf{Real-to-sim Policy Evaluation:}
A few recent works have recently explored whether increased visual and physical realism can improve the ability of simulated environments to be predictive of real-world policy performance:~\cite{chhablani2025embodiedsplatpersonalizedrealtosimtorealnavigation, escontrela2025gaussgymopensourcerealtosimframework} demonstrated that navigation policies can be evaluated in realistic reconstructions of indoor environments, and~\cite{yang2023unisim} demonstrated the same for autonomous driving policies. In robot manipulation, where policies closely \emph{interact} with their environment, the requirements for physical and visual realism of simulated evaluation environments increase: SIMPLER~\cite{li24simpler} first showed that simulation can be used for evaluation of real-robot manipulation policies, but it fails to deliver strong real-world performance correlation for recent generalist policies~\cite{kim24openvla}. Additionally, the ``green-screening'' approach used by~\cite{li24simpler} and follow-up works~\cite{jangir2025robotarenainftyscalablerobot}, which copy-pastes background pixels from real-robot images, does not support evaluation for robot setups with moving cameras. Unfortunately, this excludes most generalist policies~\cite{geminirobotics, nvidia2025gr00tn1openfoundation, pi05, blackpi0}, which feature moving wrist cameras attached to the robot's end-effector. Our work uses 3D~Gaussian splatting techniques to create high-fidelity simulation environments. The resulting environments show strong real-to-sim performance correlation \emph{and} support the evaluations of policies with wrist cameras.

\noindent\textbf{Real-to-sim Environment Creation:}
A number of works have developed methods for digitizing real-world environments. Some approaches focus solely on room layout, and fill the scene by sampling from a large 3D object database~\cite{deitke2023phone2proc}, often lacking in visual realism. For higher visual quality, early approaches relied on specialized and expensive sensory equipment~\cite{savva2019habitat, xia2018gibson}, and more recently, learned models have been used to bypass such requirements and instead infer 3D~models of environments from raw image or video inputs~\cite{torne2024rialto, escontrela2025gaussgymopensourcerealtosimframework}. In this context, Gaussian splatting has become a popular tool for inferring scene geometry and appearance from raw sensory observations ~\cite{3dgs, chen2025survey3dgaussiansplatting, 2dgs, irshad2024neuralfieldsroboticssurvey}. Multiple works have explored the use of Gaussian splatting techniques in robotics ~\cite{xia2025drawerdigitalreconstructionarticulation, yu2025pogs,chakra25realissim, yu2024language,  qureshi2025splatsim, dan2025xsim, yu2025real2render2realscalingrobotdata}. While these works have shown that splatting can produce environments that can be used for augmenting the training of robot policies, in this work we investigate the utility of real-to-sim techniques in the context of robot policy \emph{evaluation}. Most closely related to our work,~\cite{jiang2025gsworld, zhang2025real} used simulated environments created via Gaussian splatting for policy evaluation, but focused on evaluating narrow, single-task policies in environments seen in the training data. In contrast, comprehensive evaluation of modern \emph{generalist} policies requires evaluations across a large number of tasks in unseen environments~\cite{atreya2025roboarena}. Our work is the first to develop a pipeline that enables evaluation of real-robot policies in \emph{unseen} simulated environments, with strong performance correlation to real-world evaluations. 

%% file: sections/03_prelims.tex
\section{Preliminaries}
\label{sec:prelim}

The goal of this work is to evaluate \emph{real-world} generalist robot policies in \emph{simulated} environments such that the simulated evaluation faithfully reflects real-world performance. %
Let $\Pi = \{\pi_1, \dots, \pi_N\}$ denote a set of  policies, each of which can be executed both in the real world and in simulation. Let $\mathcal{E}_{\text{real}}$ denote the real-world environment distribution, and $\mathcal{E}_{\mathcal{S}}$ denote the corresponding simulated distribution constructed by a generative simulation pipeline $\mathcal{S}$.

For each policy $\pi_i \in \Pi$, we can define a metric of expected real-world performance (e.g., success rate) as $R_i = \mathbb{E}_{e \sim \mathcal{E}_{\text{real}}} \big[\,r(e, \pi_i)\,\big]$, where $r(e, \pi_i)$ denotes the episodic performance when executing policy $\pi_i$ in environment $e$. Analogously, we define the policy’s performance in simulation as $R_{\mathcal{S}, i} = \mathbb{E}_{e \sim \mathcal{E}_{\mathcal{S}}} \big[\,r(e, \pi_i)\,\big].$ The goal of simulation-based evaluation is to design simulation environments $\mathcal{E}_{\mathcal{S}}$ such that the ranking and relative performance of policies in simulation closely approximate their real-world behavior. 
We follow prior work \cite{li24simpler} to quantify the \emph{faithfulness} of policy evaluation using two metrics. 

\textbf{(i) Pearson correlation coefficient ($r$):} This measures the linear correlation between real and simulated performance:
\[
r(R, R_{\mathcal{S}}) = 
\frac{\sum_{i=1}^{N} (R_i - \bar{R})(R_{\mathcal{S}, i} - \bar{R}_{\mathcal{S}})}
{\sqrt{\sum_{i=1}^{N} (R_i - \bar{R})^2}\sqrt{\sum_{i=1}^{N} (R_{\mathcal{S}, i} - \bar{R}_{\mathcal{S}})^2}},
\]
where $\bar{R}$ and $\bar{R}_{\mathcal{S}}$ are the mean real and simulated scores.

\textbf{(ii) Mean Maximum Rank Violation (MMRV):} This captures the severity of rank inversions between real and simulated evaluations. Intuitively, it measures the average worst-case real-world performance gap between any two policies that are mis-ranked under simulation:
\begin{align}
\mathrm{MMRV}(R, R_{\mathcal{S}}) = \frac{1}{N}\sum_{i=1}^{N} \max_{1 \le j \le N}
|R_i - R_j| \cdot \mathbf{1}\!\Big[
    (R_{\mathcal{S}, i} < R_{\mathcal{S}, j}) 
    \neq 
    (R_i < R_j)
\Big]. \nonumber
\label{eq:mmrv}
\end{align}

The ideal simulation pipeline $\mathcal{S}$ has high Pearson correlation ($r \to 1$) and a low MMRV ($\to 0$). This indicates a well-calibrated simulation, in which relative performance ordering and absolute success trends of real-world policies are faithfully preserved.

\textbf{2D Gaussian Splatting for Scene Reconstruction:} 
To faithfully recover the scene geometry, we resort to recent neural reconstruction tools~\cite{3dgs, 2dgs, mildenhall2020nerf, bakedsdf, chen2025survey3dgaussiansplatting}. To achieve the best trade-off between high visual fidelity and accurate mesh geometry, we leverage 2D Gaussian Splatting (2DGS)~\cite{2dgs}, which represents a scene as a collection of oriented planar Gaussian disks embedded in 3D space, enabling improved geometric fidelity compared to other representations. A 2D Gaussian splat is represented as a tuple
$
g_j = (\boldsymbol{\mu}_j,\; \mathbf{R}_j,\; \mathbf{S}_j,\; \mathbf{c}_j,\; \alpha_j),
$
where $\boldsymbol{\mu}_j \in \mathbb{R}^3$ is the disk center, $\mathbf{R}_j \in \mathrm{SO}(3)$ encodes its orientation, $\mathbf{S}_j \in \mathbb{R}^{3\times 3}$ defines its covariance in the disk plane, and $(\mathbf{c}_j,\alpha_j)$ denote its color and opacity. A full scene is represented as a collection of these Gaussian splats $\mathcal{D} = \{g_j\}_{j=1}^{M}$, optimized so that a differentiable renderer reproduces the input views on aggregate. 
Once optimized, the resulting representation supports both real-time photorealistic rendering and accurate mesh extraction, making it well suited for physics-based simulation of robotic policies. Additional implementation details are provided in Appendix~\ref{app:2dgs-prelim}.

\textbf{Generalist Robot Policies:} While \emph{generalist} policies is generally an ill-defined term, in this work we specifically use it to refer to a class of language-conditioned multi-task robot policies known as Vision-Language-Action models (VLAs)~\cite{geminirobotics, pi05, blackpi0, nvidia2025gr00tn1openfoundation}. VLAs are built on top of transformer based Vision-Language Models (VLMs), fine-tuned to produce robot actions conditioned on visual observations and textual instructions. We denote a generalist policy as $\pi_{\theta} : (\mathcal{O}_{\text{vis}}, \mathcal{O}_{\text{lang}}) \rightarrow \mathcal{A}$, where $\mathcal{O}_{\text{vis}}$ is the space of observations, $\mathcal{O}_{\text{lang}}$ is the space of language instructions, and $\mathcal{A}$ is the action space. 

%% file: sections/04_methods.tex
\section{Scalable Simulation Evaluations with Real-World Video Scans}
\label{sec:approach}

\begin{figure*}
    \centering
    \includegraphics[width=\linewidth]{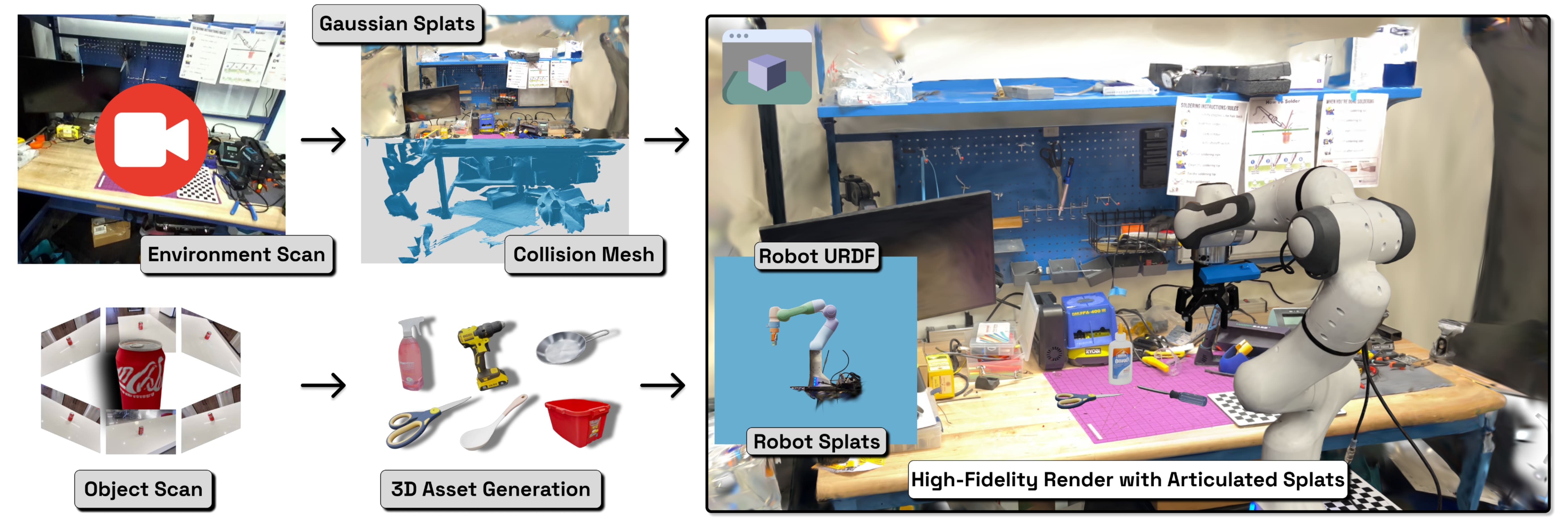}
    \caption{\footnotesize{\textbf{Environment creation in \Acronym{}.} A user first scans in a real-world environment, robot and objects. They then use \Acronym{} to create articulated Gaussian splat representations of the real-world environment that capture geometry and visual appearance of the environment. Finally, they compose the evaluation scene by combining scanned-in scene, objects and robot to use for policy evaluation.}}
    \label{fig:overview}
\end{figure*}

In this section, we describe \Acronym{}, our real-to-sim system for scalable creation of high-fidelity simulated environments (\Cref{fig:overview}). We structure our discussion around two components: first, we detail the construction pipeline that transforms a video recording of the real-world scene into an interactive simulation using 2DGS and automatic articulation; second, we address remaining real-to-sim discrepancies by proposing a lightweight simulation data co-training strategy to ensure well-correlated evaluations. Together, these components enable~\textit{zero-shot} evaluation of real-world generalist policies in previously unseen simulation environments.

\subsection{Environment Generation with \Acronym{}}
\label{sec:env_generation}

To create an environment with \Acronym{}, we take a compositional approach. We first construct the visuals and geometry of the environments, then construct the visuals, geometry, and articulation of the robot, and finally add object assets into the scene. We outline these steps below:

\noindent\textbf{Environment Creation via Reconstruction:} We start by capturing a short (2--5~minutes) monocular video recording of the scene. We then run a reconstruction pipeline based on 2DGS~\cite{2dgs} to reconstruct scene visuals and geometry. To facilitate compositional scene construction, we record separate video sequences for (i)~the static background scene and (ii)~the robot. This enables training of distinct Gaussian splat models for each component, which can later be composed and articulated in Euclidean space. During capture, the user slowly moves the camera to avoid motion blur while ensuring full coverage of all visible surfaces. To assist in automatically resolving global orientation and metric scale, a ChArUco board is placed in the environment (see \Cref{fig:overview}). This data acquisition process for a scene typically requires only 5--10~minutes. Following data capture, we estimate the camera parameters for all frames using COLMAP~\cite{schonberger2016colmap} and train a 2D~Gaussian Splatting model $\mathcal{D} = \{g_j\}_{j=1}^{M}; g_j = (\boldsymbol{\mu}_j,\; \mathbf{R}_j,\; \mathbf{S}_j,\; \mathbf{c}_j,\; \alpha_j)$ for each recording by optimizing a standard photometric reconstruction loss as described in Appendix~\ref{app:2dgs-prelim}.

To enable closed-loop robotic evaluation in a physics-based simulator, we require the geometric and photometric components of the reconstructed scene. We obtain a Truncated Signed Distance Field (TSDF) using a combination of rasterizing the splats and volumetric depth fusion \cite{curless1996tsdf}. An explicit mesh representation $\mathcal{M}$ is then extracted from the TSDF using marching cubes~\cite{LorensenCline87}. The resulting meshes are imported into IsaacSim~\citep{NVIDIA_Isaac_Sim} to enable realistic contact dynamics and object interactions, while the underlying Gaussian splats $\mathcal{D}$ provides the corresponding photometric renders for high-fidelity visual affects. The environment creation framework can be fully automated and typically completes within 30~minutes on an NVIDIA RTX~4090. Please refer to Appendix~\ref{app:2dgs-method} for implementation details.

\noindent\textbf{Automatic Articulation:}
While the environment remains static, the robot mesh and the corresponding Gaussian splats must be \emph{articulated} for interactive evaluation. To enable this, we anchor subsets of Gaussian splat primitives to the robot’s kinematic links, allowing them to move according to the robot’s joint configuration. Let the robot have links $\{\mathcal{L}_1, \ldots, \mathcal{L}_L\}$ with joint configuration $\mathbf{q} \in \mathbb{R}^d$. Each link $\mathcal{L}_\ell$ is associated with a set of Gaussian primitives $\mathcal{G}_\ell = \{g_{\ell j}\}_{j=1}^{M_\ell}$ and a rigid-body transform $T_\ell(\mathbf{q}) = [R_\ell(\mathbf{q}) \mid \mathbf{t}_\ell(\mathbf{q})] \in \mathrm{SE}(3)$ obtained via forward kinematics. For a configuration $\mathbf{q}$, the position and covariance of each Gaussian is updated with the corresponding transform:$
\boldsymbol{\mu}'_{\ell j} = R_\ell(\mathbf{q})\,\boldsymbol{\mu}_{\ell j} + \mathbf{t}_\ell(\mathbf{q}), 
$ $
\Sigma'_{\ell j} = R_\ell(\mathbf{q})\,\Sigma_{\ell j}\,R_\ell(\mathbf{q})^\top.$
The articulated robot splat is then given by $\mathcal{G}(\mathbf{q}) = \bigcup_{\ell=1}^{L} \{(\boldsymbol{\mu}'_{\ell j}, \Sigma'_{\ell j}, \mathbf{c}_{\ell j}, \alpha_{\ell j})\}_{j=1}^{M_\ell}$, which can be rendered using the standard differentiable 2DGS pipeline. This enables interactive, photorealistic robots whose Gaussian splat components follow the true physical kinematics of the underlying mesh model. This allows the evaluation to be fully interactive in closed loop, while retaining the visual realism of Gaussian splatting. The fully scanned and articulated robot model can be reused across simulation environments, making this a one-time process per robot system.

\noindent\textbf{Object Creation with Generative Models:} We find that directly extracting high-quality, watertight assets from real-world scenes is a time-consuming and tedious process that often requires depth-enabled hardware and precise registration tools. Although \Acronym{} is not tied to any specific object reconstruction methods, we employ the TRELLIS 3D image-to-3D generative model~\citep{xiang2025structured} for simplicity and flexibility. TRELLIS introduces a structured representation that enables mapping from a single or a few multiview 2D images to diverse 3D representations (such as Gaussian splats, radiance fields, and meshes) with high fidelity, while recovering unobservable object regions.
With multiview images captured with a standard camera as inputs, we segment the object using SAM2~\cite{ravi25sam2} and feed the segmented views into TRELLIS to obtain a structured latent representation. This latent is decoded into a Gaussian splat for appearance and an untextured mesh for geometry, after which appearance is baked onto the mesh to produce a fully textured object for simulation.

\begin{figure}
    \centering
    \includegraphics[width=\linewidth]{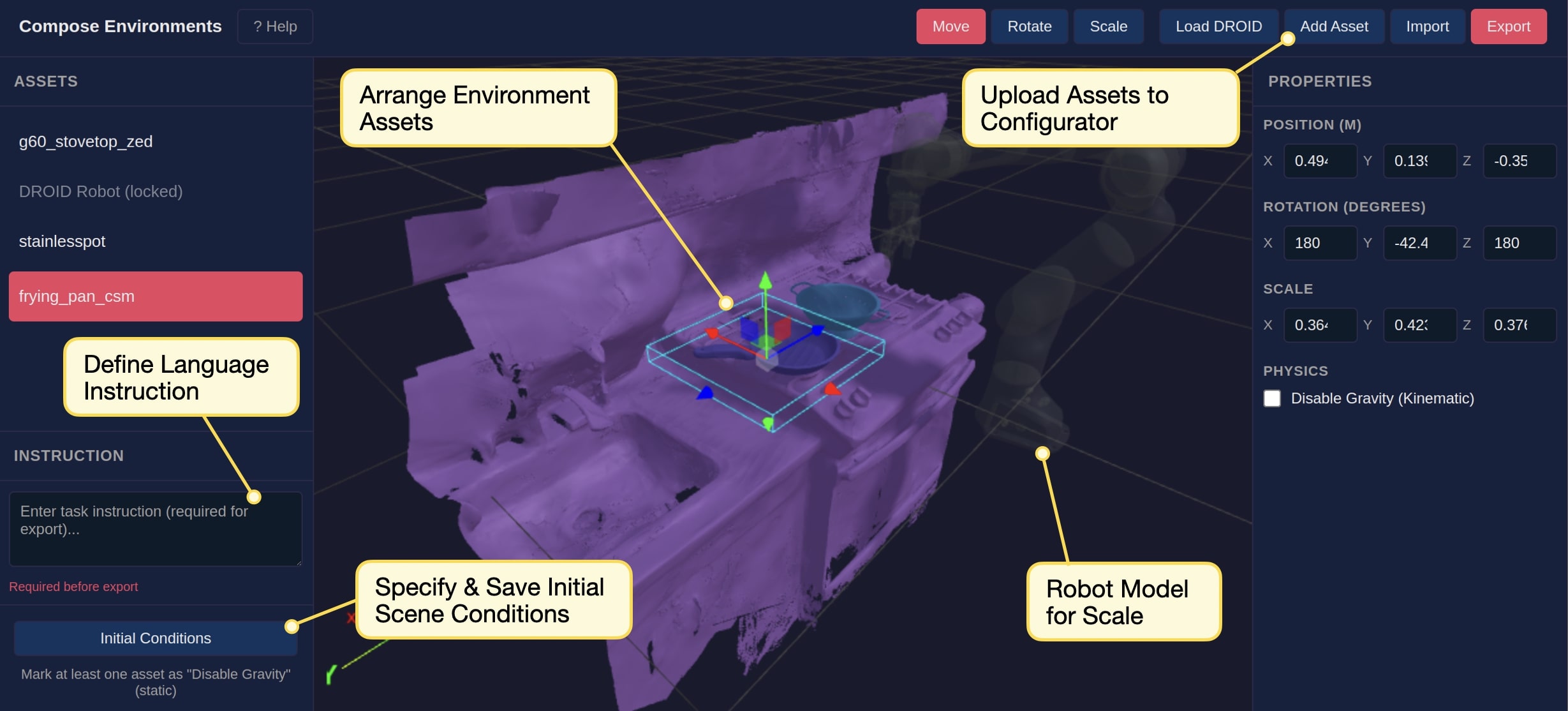}
    \caption{\footnotesize{Example of a scene being composed in our \href{https://polaris-evals.github.io/compose-environments/}{scene composition GUI}. Users can easily import environment and object assets into the tool, automatically load the DROID robot, and save the simulation ready environment into a USD. The procedure typically takes less than 5 minutes.}}
    \label{fig:interface-viz}
\end{figure}
\noindent\textbf{Composing simulated evaluation scenes.}
The final step is to assemble these components into an interactive evaluation scene using our Scene Composition GUI (\Cref{fig:interface-viz}). Since all components are represented as Gaussian primitives, the environment, robot and objects can be composed and transformed in Euclidean space. Moreover, we set standard physics parameters for contact dynamics using default values from IsaacSim, and use manufacturer-provided defaults for the robot arm. Object mass parameters can be estimated by the users. Once the scene is assembled, it can be exported as a USD file for evaluation. Alongside scene composition, the user also defines a success criterion for scoring evaluation rollouts -- e.g.,  whether a particular object is placed within a certain receptacle. %
The total human effort required for creating a new evaluation environment is typically \textbf{less than 20~minutes}, and the total wall time for creating a new evaluation environment, including training time of the Gaussian splats, is typically \textbf{less than one hour}. This is more scalable than prior simulated evaluation works~\cite{li24simpler}, which require hand-defined construction of the environment geometry and extensive tuning of lighting and textures.

An important facet of the constructed simulation is the ability to provide policies with rich sensor inputs. As opposed to prior work~\cite{li24simpler, jangir2025robotarenainftyscalablerobot} which copied image backgrounds in 2D via green-screen, \Acronym{} reconstructs the entire scene in 3D, providing full 3D rendering and novel view synthesis. This allows us to easily render third person views from any angle and provide wrist camera inputs to the policy. This opens up a new class of simulation evaluations for generalist policies that prior real-to-sim evaluation frameworks that rely on background inpainting~\cite{li24simpler, jangir2025robotarenainftyscalablerobot} were unable to support. 

\subsection{Enabling High Correlation Policy Evaluation with \Acronym{} using Co-Training}
\label{sec:cotraining}

\begin{figure}
    \centering
    \includegraphics[width=\textwidth]{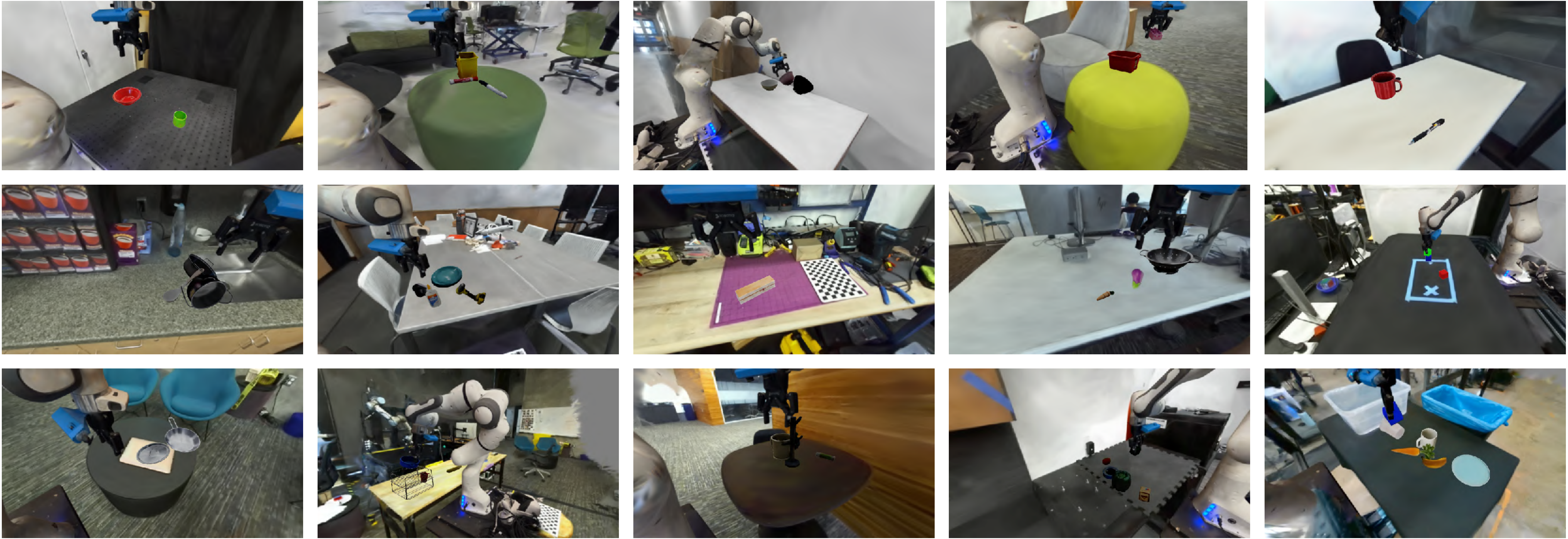}
    \caption{\footnotesize{\textbf{\Acronym{} simulated co-training dataset environments.} We collect a small number of demonstrations in simulation and co-finetune policies with this data for improved real-to-sim evaluation correlation. Importantly, once finetuned, policies evaluated in \Acronym{} show strong real-to-sim correlation, even in \emph{unseen} environments. New simulated evaluation environments are easily added without needing to collect additional demonstrations.
    }}
    \label{fig:sim_cotraining_data}
\end{figure} 

While the pipeline outlined above allows for the creation of high-fidelity environments, our empirical validation in \Cref{sec:experiments} shows that the resulting environments do not reliably lead to high real-to-sim correlation out-of-the-box. We hypothesize, that this mainly stems from subtle domain gaps between the real world and simulation, such as minor lighting variations or missing inter-object shadows. Although these differences may appear negligible to human observers, they make a considerable difference in policy performance. To better understand this effect, we conducted hardware-in-the-loop experiments that use real-world dynamics but simulated visuals, which indicate that visual mismatch is indeed the dominant source of error (see \Cref{sec:app:tasks_and_hyperparameters} for details). To mitigate these gaps and improve real-to-sim correlation, we propose a simple, policy-agnostic simulation data co-training scheme. 

Concretely, we collect a small simulation dataset via human teleoperation:
$
\mathcal{D}_{\mathcal{S}} = \{ (o_t^{\mathcal{S}}, a_t^{\mathcal{S}}, p_t^{\mathcal{S}}) \}_{t=1}^{M},
$
consisting of simulated visual observations $o_t^{\mathcal{S}}$, expert actions $a_t^{\mathcal{S}}$, and proprioceptive states $p_t^{\mathcal{S}}$, gathered over a set of simulated co-training environments $\mathcal{E}_{\mathcal{S}}^{\text{train}} \subset \mathcal{E}_{\mathcal{S}}$. Given a pretrained generalist policy $\pi_\theta$ with original training data $\mathcal{D}_{\text{pre}}$, we update the policy parameters via \emph{co-training} by sampling batches that mix simulation data with the original pretraining data:
\[
\theta' \leftarrow 
\theta - \eta \nabla_\theta \,
\mathbb{E}_{(o,a,p) \sim (1-\lambda)\mathcal{D}_{\text{pre}} + \lambda \mathcal{D}_{\mathcal{S}}}
\big[\,\mathcal{L}_{\text{BC}}(\pi_\theta(o,p), a)\,\big],
\]
where $\mathcal{L}_{\text{BC}}$ is a standard behavior cloning objective and $\lambda \in [0,1]$ controls the ratio of simulation to pretraining samples in each batch. Intuitively, co-training with simulation data improves the policy’s robustness to subtle visual gaps between simulation and the real world, leading to more reliable policy evaluation in simulation. In practice, we find that only a few hundred simulated episodes and a few hundred fine-tuning steps are sufficient to substantially improve real-to-sim correlation, typically completed in $<25$~mins. This is orders of magnitude less data and compute than what is used during generalist pretraining. Importantly, because simulation co-training aims to refine visual representations instead of acquire new task-specific skills, the simulated demonstrations \emph{are not required} to be collected in the target simulation environment(s) or task(s) $\mathcal{E}_{\mathcal{S}}^{\text{test}}$. Instead, they can be collected apriori, on a completely separate set of environments and tasks - in our experiments co-training environments ($\mathcal{E}_{\mathcal{S}}^{\text{train}}$) and evaluation environments ($\mathcal{E}_{\mathcal{S}}^{\text{test}}$) are disjoint. As a result, the co-trained policies can be evaluated \emph{zero-shot} in unseen simulation environments.

%% file: sections/05_experiments.tex
\section{Experiments}
\label{sec:experiments}

    Our experimental analysis aims to answer the following questions: (1)~Does \Acronym{} enable zero-shot evaluation in \emph{unseen} simulation environments with strong correlation to real-world performance? (2)~Which design decisions affect zero-shot real-to-sim correlation? (3)~Does \Acronym{} provide stronger real-to-sim evaluation correlation than current simulated evaluation tools in robotics?

\begin{figure}
    \centering
    \includegraphics[width=\linewidth]{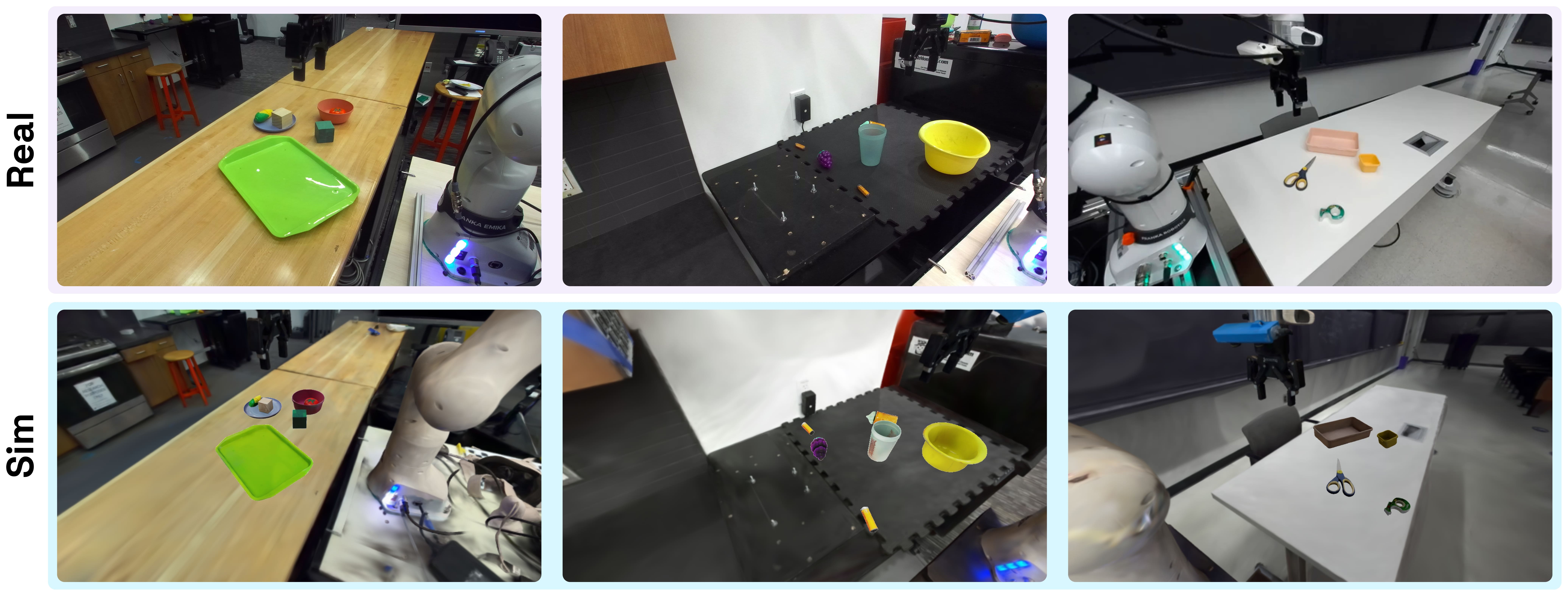}
    \caption{\footnotesize{\textbf{Evaluation environments.} \textbf{Top}: real-world environments, \textbf{Bottom}: \Acronym{} simulated evaluation replicas. We create high visual fidelity environments utilizing Gaussian Splats for environment reconstruction and TRELLIS for object asset generation. 
    }}
    \label{fig:real_vs_sim_envs}
\end{figure}

\subsection{Experimental Setup}

\textbf{Robot platform.} We perform our experiments using the DROID~\citep{khazatsky24droid} platform--one of the mostly widely used robot platforms for generalist policies~\citep{pertsch2025fast, pi05, atreya2025roboarena}--which consists of a 7-DoF Franka Panda robot arm, and a set of two ZED RGB cameras with one mounted to the robot's wrist. While DROID offers both joint velocity and joint position action spaces, we focus on evaluating policies using joint \emph{position} actions, which we found easier to simulate accurately since they suffer from less accumulating error.

\textbf{Policies.} We use a representative set of DROID VLAs from prior works for testing the efficacy of \Acronym{} as an evaluation tool: (1) $\bm{\pi_0}$~\citep{blackpi0} and $\bm{\pi_0}$\textbf{-FAST}~\citep{pertsch2025fast}, which are diffusion-based and auto-regressive VLAs respectively, (2) \textbf{PaliGemma-binning}, an auto-regressive VLA with binning action tokenization from~\cite{pertsch2025fast}, and (3) $\bm{\pi_{0.5}}$~\cite{pi05}, the current state-of-the-art VLA on DROID~\cite{atreya2025roboarena}.

\textbf{Tasks.} To enable precise evaluation of real-to-sim correlations, we set up 6 \emph{paired} real-world and simulated environments across two institutions (showcases in \Cref{fig:real_vs_sim_envs}). The environments span diverse visual scenes and a representative sample of rigid object manipulation tasks. We also create an additional 15~real-to-sim scenes through our pipeline, which we use to collect a dataset of around 350~simulation demonstrations for co-training (\Cref{sec:cotraining}). We ensure that these ``co-training environments" are \emph{completely separate from our evaluation environments and tasks} and none of the scenes or objects are shared.

\textbf{\Acronym{} evaluation procedure.} We finetune all policies for 1k~steps on the co-training dataset using supervised learning, then run 50~rollouts per task in simulation to compute the policy's performance. Each task is automatically scored on a progress scale from $[0 \dots 1]$. For details on all tasks, progress scales, and hyperparameters, please see Appendix \ref{sec:app:tasks_and_hyperparameters}.

\begin{figure}[t]
\centering

\newlength{\CardImgH}
\setlength{\CardImgH}{0.34\linewidth}

\begin{minipage}[t]{0.495\linewidth}
\centering
\begin{tcolorbox}[
  enhanced,
  colback=polaris-bg-elevated,
  colframe=polaris-border-subtle,
  arc=3mm,
  boxrule=0.4pt,
  left=2pt,right=2pt,top=2pt,bottom=2pt,
  equal height group=polarisrow, %
]
\parbox[c][\CardImgH][c]{\linewidth}{%
  \centering
  \includegraphics[
    trim={10pt 8pt 10pt 6pt},clip,
    width=\linewidth,height=\CardImgH,keepaspectratio
  ]{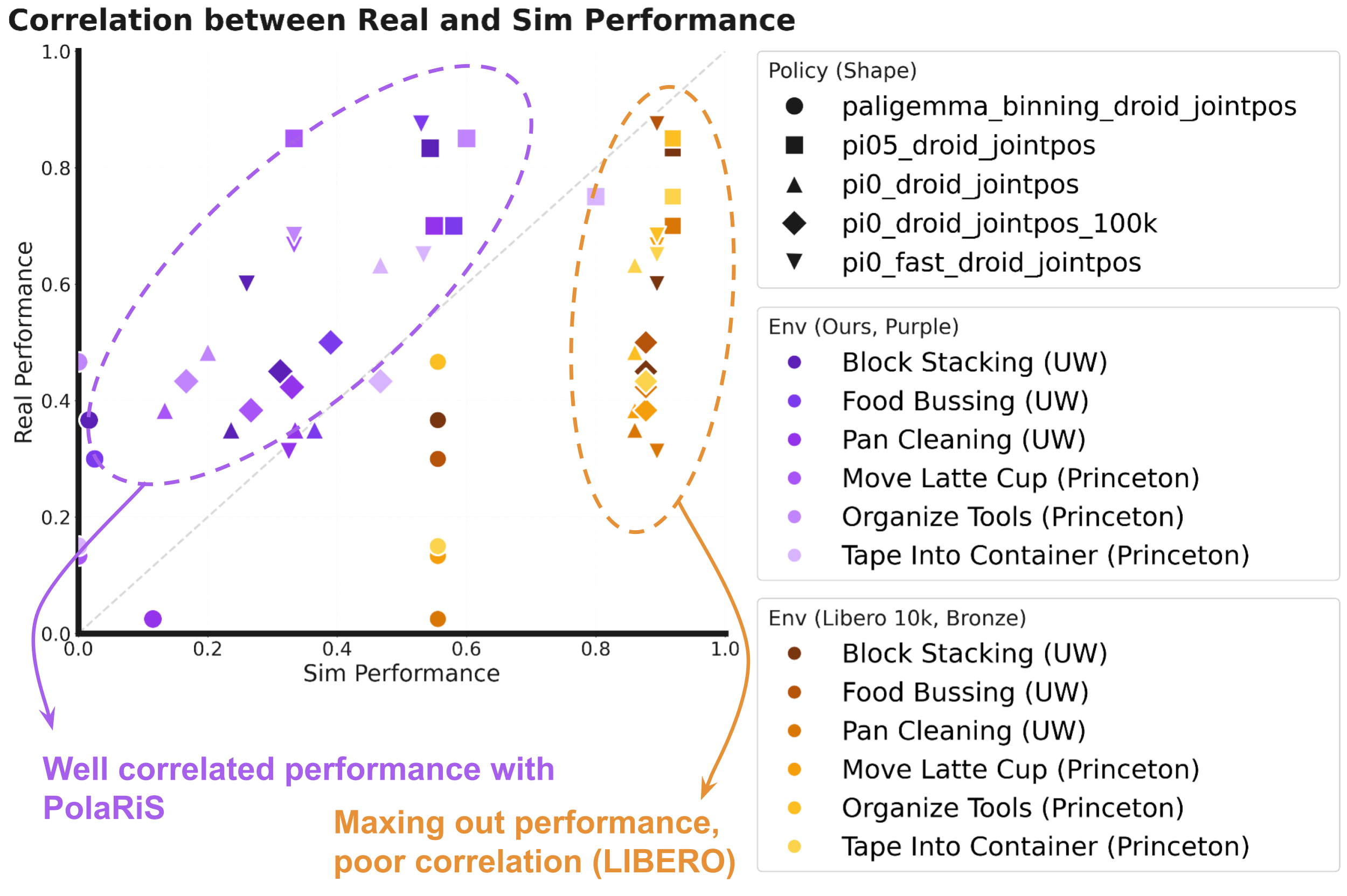}%
}

\vspace{1pt}
\vfill %

\captionsetup{type=figure,skip=0pt,aboveskip=0pt,belowskip=0pt}
\caption{\footnotesize\textbf{Correlation of Evaluations between Real and Sim:}
\Acronym{} achieves strong real-to-sim correlation in \emph{unseen} environments,
accurately ranking real-world policy performance via simulation.}
\label{fig:correlation}
\end{tcolorbox}
\end{minipage}
\hfill
\begin{minipage}[t]{0.495\linewidth}
\centering
\begin{tcolorbox}[
  enhanced,
  colback=polaris-bg-elevated,
  colframe=polaris-border-subtle,
  arc=3mm,
  boxrule=0.4pt,
  left=2pt,right=2pt,top=2pt,bottom=2pt,
  equal height group=polarisrow,
]
\parbox[c][\CardImgH][c]{\linewidth}{%
  \centering
  \includegraphics[
    trim={10pt 5pt 10pt 0pt},clip,
    width=\linewidth,height=\CardImgH,keepaspectratio
  ]{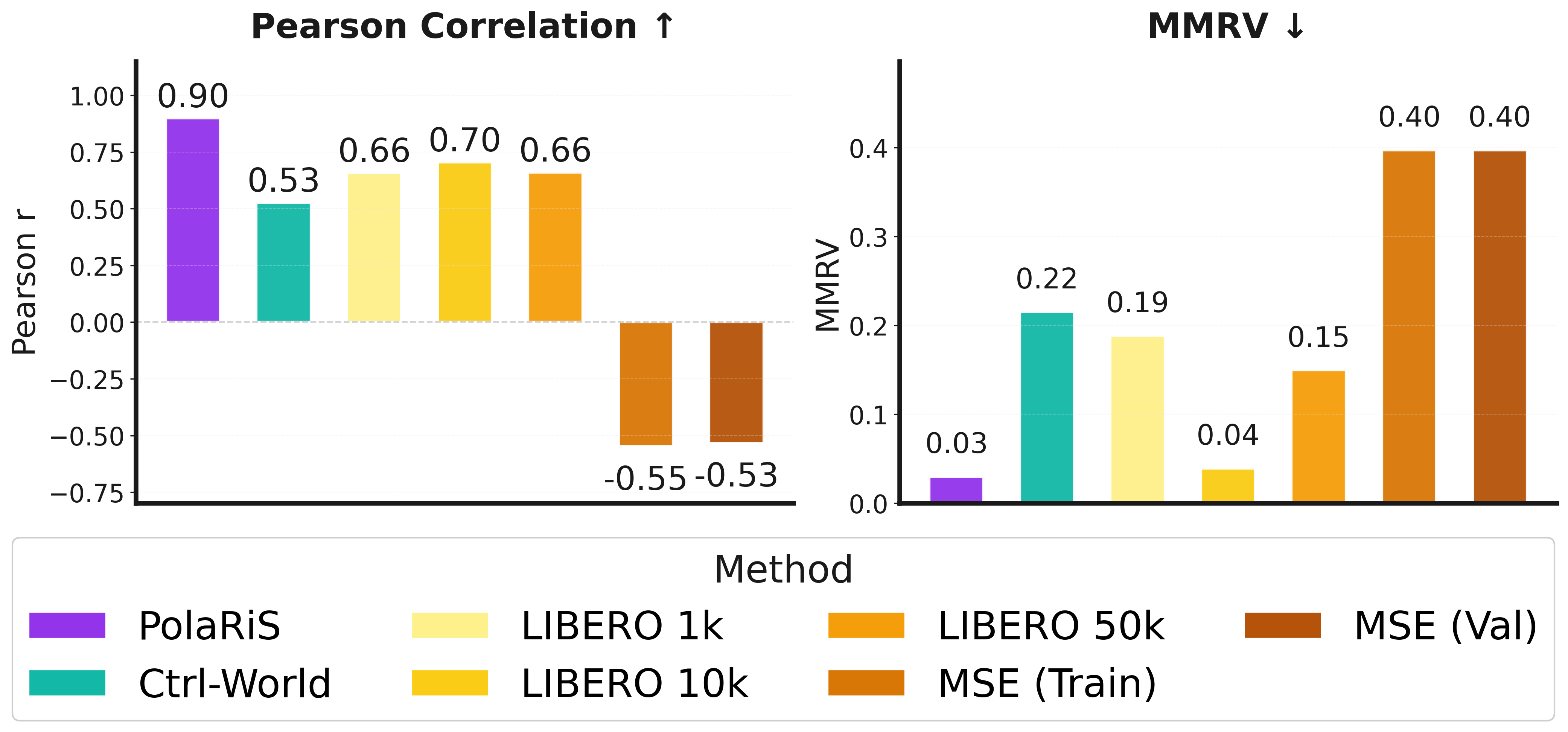}%
}

\vspace{1pt}
\vfill

\captionsetup{type=figure,skip=0pt,aboveskip=0pt,belowskip=0pt}
\caption{\footnotesize\textbf{Comparison of Evaluation Metrics:}
We compare Pearson correlation coefficient and Mean Maximum Rank Violation (MMRV).
\Acronym{} achieves the best performance.}
\label{fig:correlation_barplots}
\end{tcolorbox}
\end{minipage}

\end{figure}

\textbf{Comparisons.} We compare our real-to-sim evaluation approach to existing methods used for offline policy evaluation: \textbf{MSE}, which computes training / validation set action MSE for each policy; \textbf{Libero-Score}, which evaluates policies on the Libero simulation benchmark~\citep{liu23libero} that is often used in VLA benchmarking~\citep{huang25otter, kim25vla, kim24openvla, lee2025molmoactactionreasoningmodels}; and \textbf{Video model evaluation}, which rolls out policies in the open-source \textbf{Ctrl-World}~\citep{guo2025ctrl} video model and uses humans to score the rollouts. For Libero, to get the highest-quality evaluations possible, we finetune all policies on Libero-90, the largest of the Libero datasets, for 50k~steps and evaluate on all 90~tasks, running 50~rollouts per task or a total of 4,500~evaluations per policy. Notably, this is a more thorough evaluation than done by most existing VLA papers using Libero, which typically evaluate on much smaller Libero splits~\citep{huang25otter, kim25vla, kim24openvla}. Additionally, we evaluate multiple Libero-finetuned checkpoints (1k, 10k, 50k) and report results for the best-correlated checkpoint in \Cref{fig:correlation}. As a result, the Libero-Score approach uses approximately 10x more finetuning and evaluation compute than our simulated evaluation approach. We note that existing real-to-sim evaluation approaches~\cite{li24simpler, jangir2025robotarenainftyscalablerobot}, including SIMPLER, do not support evaluation of policies with wrist cameras, and thus cannot even be used to evaluate most state of the art VLA models.

\textbf{Metrics.} For each environment we run 20~real-world evaluation rollouts per policy to establish a ground-truth ranking of policy performance. As described in \Cref{sec:prelim}, we report Pearson correlation $r$ and MMRV between the real-world and simulation performance.

\subsection{Real-to-Sim Correlation in Unseen Environments}
\label{sec:main_results}

\begin{wrapfigure}{r}{0.45\linewidth}
    \centering
    \vspace{-0.4cm}
    \begin{tcolorbox}[
    enhanced,
    width=\linewidth,
    colback=polaris-bg-elevated,
    colframe=polaris-border-subtle,
    arc=3mm,
    boxrule=0.4pt,
    left=4pt,
    right=4pt,
    top=6pt,
    bottom=6pt,
]
    \includegraphics[width=0.99\linewidth]{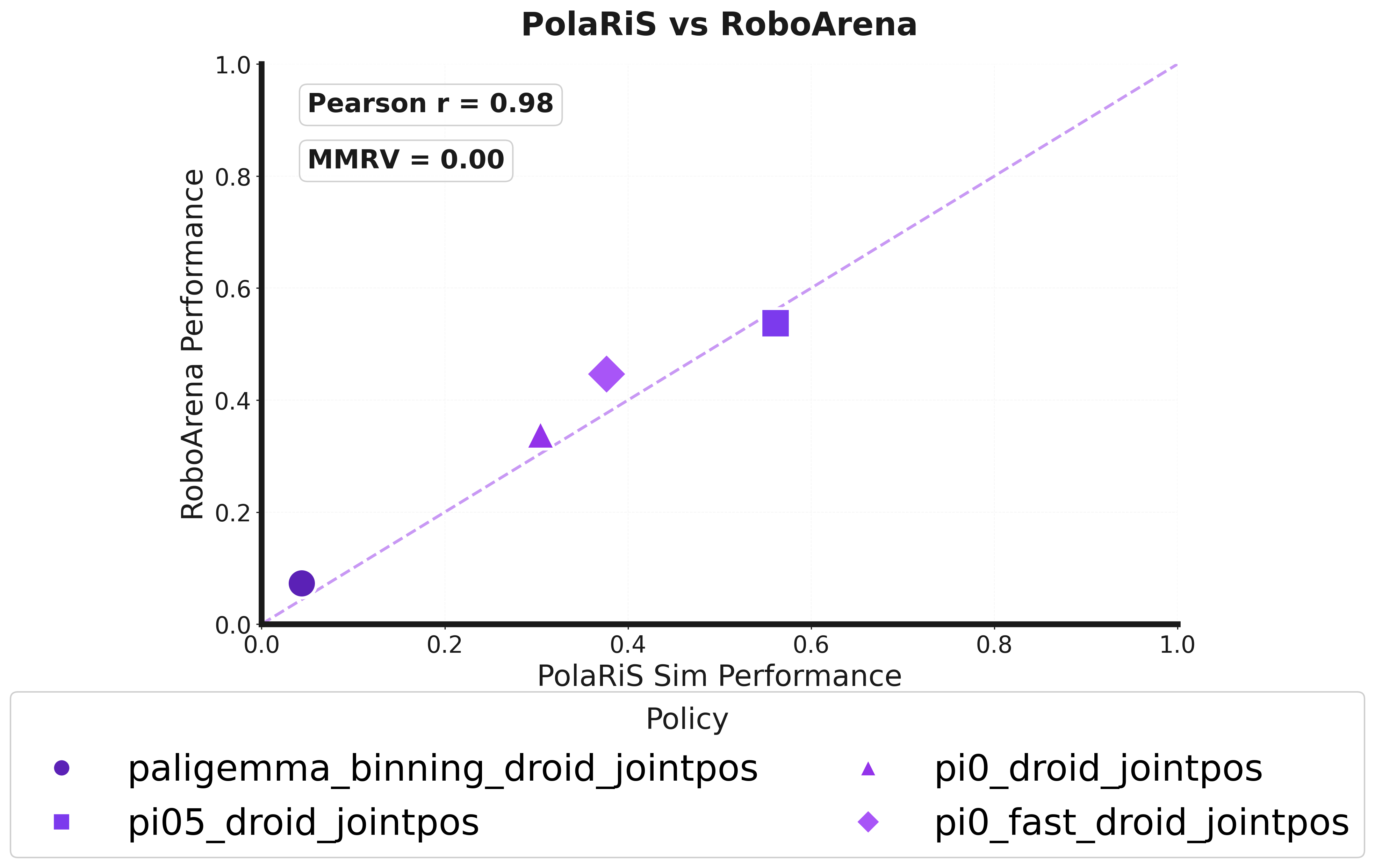}
\caption{\footnotesize{\textbf{Correlation with RoboArena~\cite{atreya2025roboarena}:} \Acronym{} evaluations show strong performance correlation with RoboArena~\cite{atreya2025roboarena} rankings.}}
    \label{fig:roboarena_results}
    \end{tcolorbox}
    \vspace{-0.5cm}
\end{wrapfigure}

We report real-world correlation results for all tested approaches in \Cref{fig:correlation} and \Cref{fig:correlation_barplots}. In line with existing literature~\cite{li24simpler}, we find that MSE is a poor metric for real-world policy performance and shows low correlation. Libero-score too is poorly correlated with real-world performance, with nearly all tested policies scoring between 90 and 95\% success in Libero, but spanning the full spectrum from high-performing to low-performing policies in the real world. %
Finally, we find that today's open-source video models are not yet reliable enough for policy evaluation in diverse scenes: in our experiments, Ctrl-World often produces heavy hallucinations during object interaction, making accurate scoring challenging and ultimately leading to clear policy mis-rankings (MMRV: 0.22). While recent works have shown promising video model evaluation results in narrower scene settings~\citep{guo2025ctrl, veorobotics2025}, our experiments suggest that more advances are needed for these models to be widely applicable tools for robot policy evaluation in diverse real-world tasks.

In contrast, \Acronym{} achieves strong correlation results across all environments: in our experiments, differences in \Acronym{} scores closely reflect real-world performance differences, and rank the performance of policies accurately. Importantly, this result does not only hold in terms of \emph{average} performance, but holds \emph{for each individual environment}: the \emph{worst-case} correlation across all six tested environments is $r=0.81$. See individual environment results in \Cref{fig:individual-correlation}. In contrast to hand-designed sim benchmarks, \Acronym{}'s pipeline can be used to estimate both: average performance of generalist policies and performance in specific deployment scenarios.

\noindent Furthermore, we find that \Acronym{} scores exhibit strong correlation to RoboArena~\citep{atreya2025roboarena} real-world evaluation scores (see \Cref{fig:roboarena_results}). We report correlation to the average progress scores reported as part of the RoboArena evaluations. While we find that scores correlate well, we emphasize that the simulated evaluations in our experiments only capture a small subset of the capabilities tested in RoboArena---particularly those involving non-rigid body dynamics, soft-body interactions, and complex contact behaviors that remain challenging for current simulators. Thus, we believe that \Acronym{} can serve as a useful tool for cheaply and rapidly estimating VLA performance during research and development, but \emph{does not} entirely replace real-world evaluation.

\subsection{Ablation Studies and Analysis}
In this section, we investigate the importance of different components of our \Acronym{} framework.

\begin{figure*}[!h]
\centering

\begin{tcolorbox}[
    enhanced,
    width=1.0\linewidth,
    colback=polaris-bg-elevated,
    colframe=polaris-border-subtle,
    arc=3mm,
    boxrule=0.4pt,
    left=4pt,
    right=4pt,
    top=6pt,
    bottom=6pt,
]

\begin{minipage}[t]{0.32\linewidth}
    \centering
    \includegraphics[height=3.5cm]{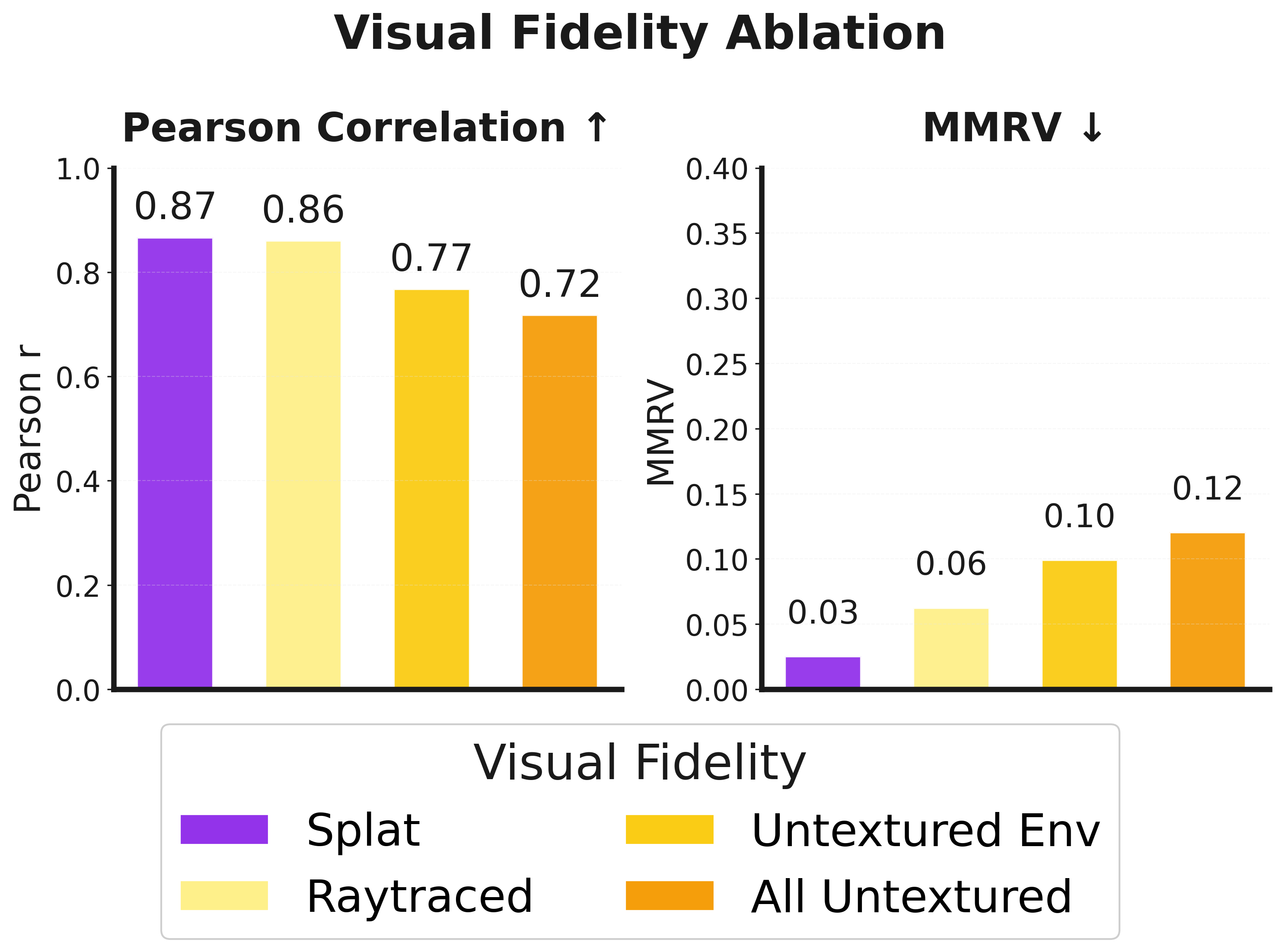}
    \caption{\footnotesize{\textbf{Visual fidelity ablation:} High-fidelity rendering is key for strong real-to-sim correlation: evaluations with texture-less simulation environments or raytraced rendering underperform Gaussian splatting based rendering evaluations, even after simulation data co-training.}}
    \label{fig:visual_fidelity}
\end{minipage}\hfill
\begin{minipage}[t]{0.32\linewidth}
    \centering
    \includegraphics[height=3.5cm]{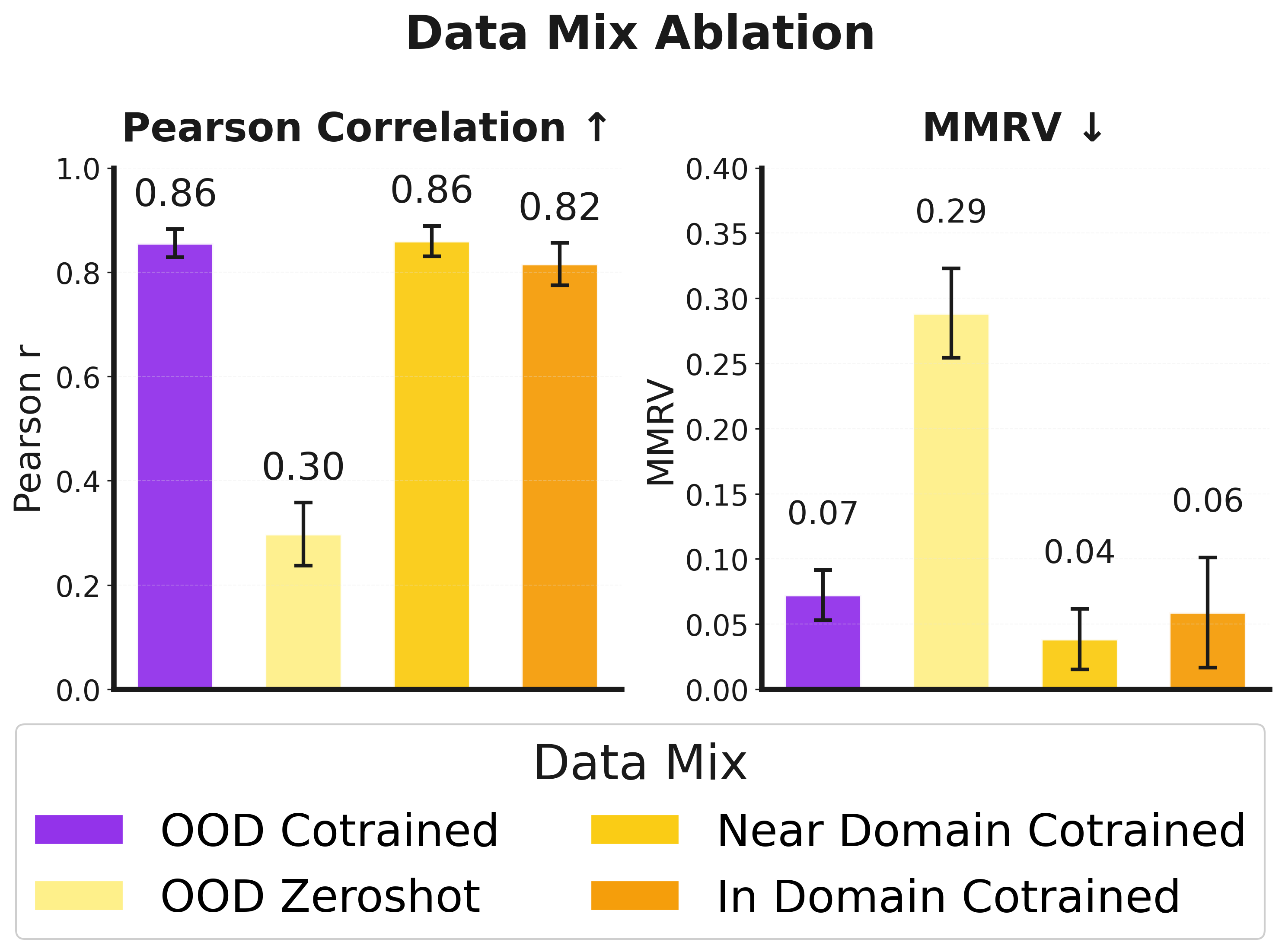}
    \caption{ \footnotesize{\textbf{Co-training dataset ablation:} Simulation data co-training is key for strong real-to-sim evaluation correlation, but importantly, once co-trained with our simulated demonstration dataset, \Acronym{} provides strong correlation even for unseen (``OOD'') simulation environments.}}
    \label{fig:cotraining_data}
\end{minipage}\hfill
\begin{minipage}[t]{0.32\linewidth}
    \centering
    \includegraphics[height=3.3cm]{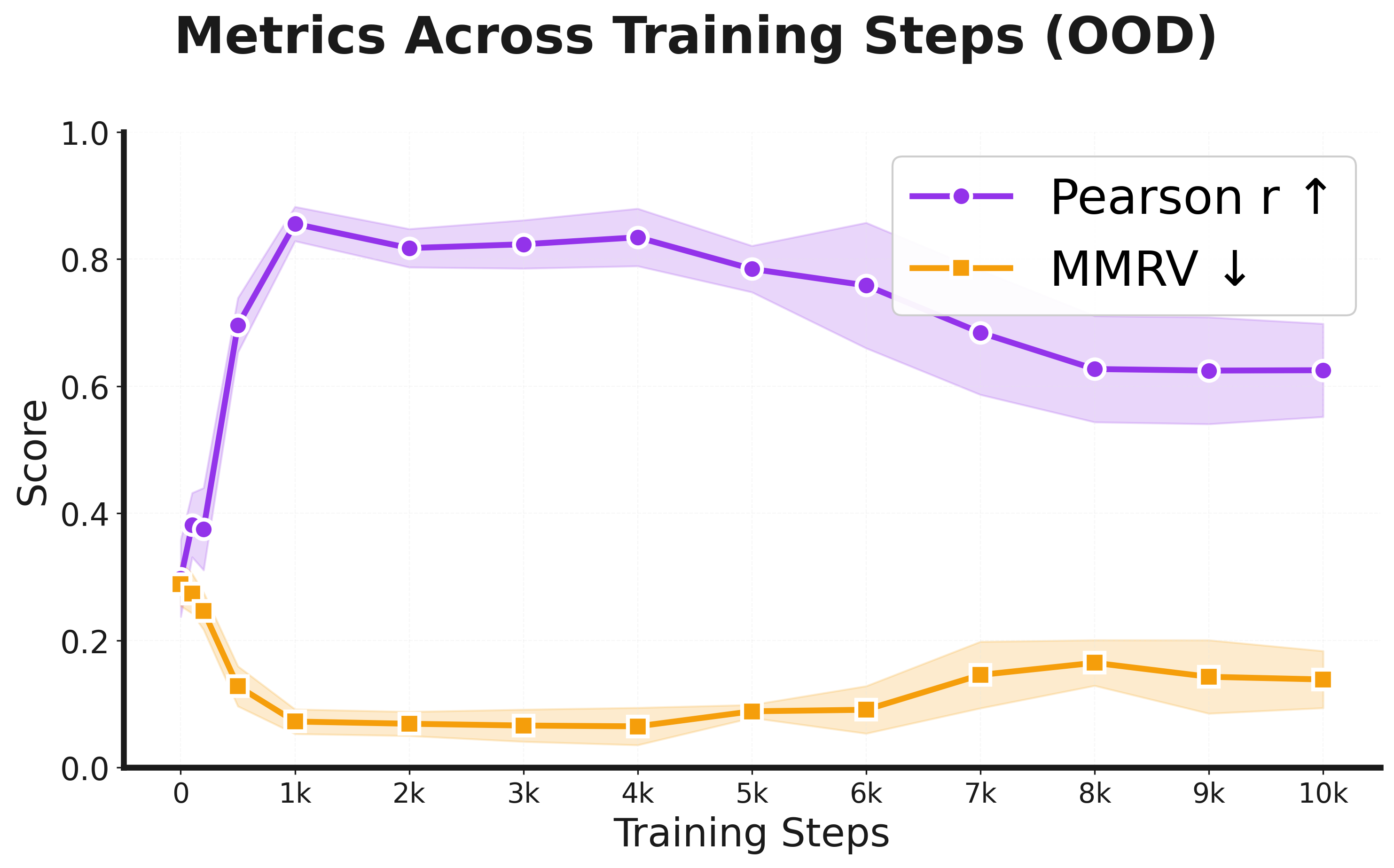}
    \caption{\footnotesize{\textbf{Finetuning steps ablation:} Real-to-sim correlation across five finetuning seeds. \Acronym{} reliably achieves strong real-to-sim correlation after just a few hundred steps of simulation data co-finetuning. Finetuning policies for too long reduces real-to-sim correlation.}}
    \label{fig:finetuning_steps}
\end{minipage}

\end{tcolorbox}

\end{figure*}

\textbf{Visual fidelity.} To ablate the impact of the visual fidelity of simulated environments on real-to-sim correlation, we create versions of our simulated evaluation scenes with different levels of visual quality (see Table \ref{table:visual_fidelity}), using ray-tracing instead of Gaussian splatting, and removing background and foreground textures. We report real-to-sim correlation after co-training all policies with the respective simulated dataset in \Cref{fig:visual_fidelity}, left. First, we find that after co-training, all simulation variants can provide decently correlated results, suggesting that the simulation data co-training can alleviate some visual sensitivity observed in prior works~\citep{li24simpler}. Yet, we find that \Acronym{} with Gaussian splatting rendering leads to the strongest correlation. While these results suggest visual fidelity is not the \emph{sole} driver of high-quality simulated evaluations, the comparison to Libero-Score (\Cref{sec:main_results}) underlines the benefit of real-to-sim evaluation techniques compared to hand-crafting low-fidelity simulated benchmarks: real-to-sim evaluation scans \emph{real-world} environments and thus allows evaluation in environments that are closer to the training distribution. 

\textbf{Co-training dataset.} We evaluate the impact of different simulation co-training data mixtures. In our original experiments, we collected data in 15~held-out simulation environments. For this experiment, we additionally collect data in the 3~target environments, on both, the target tasks and non-target tasks. We then compare 3 settings: "\textbf{OOD}" uses only data from the 15~held-out environments (the setting from the previous sections); "\textbf{near-domain}" additionally uses data from \emph{non-target tasks}, collected in the \emph{target environments}; "\textbf{in-domain}" combines the data from the held-out environments with data collected on the \emph{target tasks} in the \emph{target environments}. We also compare to directly evaluating the pre-trained checkpoints without any simulation data co-training (``\textbf{no-cotrain}'') (see Table \ref{table:data-mixes}). We report real-to-sim correlation in \Cref{fig:cotraining_data}. First, we find that the base VLA models (``no-cotrain'') achieve non-trivial performance, but correlation is too low to accurately rank policy performance, underlining the importance of simulation co-training. Comparing the different simulation data mixes, we find that the ``OOD'' data mix (ours) leads to approximately comparable correlations to in-domain data. This underlines that the function of the simulation data is primarily to bridge subtle visual gaps between simulation and real-world environments, for which OOD scenes are sufficient. Notably, this result make \Acronym{} evaluations very scalable: we can add new evaluation environments \emph{without} collecting targeted simulation data and \emph{without} needing to finetune generalist policies for every new scenario. Such "out of the box" evaluation in unseen simulation environments makes sharing and accumulating of targeted simulated evaluations across the community feasible, like it is common-place e.g. for LLM research today. Lastly, we find that ``near-domain'' data leads to slightly stronger correlations than target task demonstrations, possibly because the latter lead to quick overfitting.

\textbf{Fine-tuning steps.} We evaluate the robustness of our pipeline to different number of finetuning steps. Concretely, we run five finetuning runs for all policies, training for 10k steps each. We then evaluate the mean and variance of Pearson correlation and MMRV across all checkpoints (see \Cref{fig:finetuning_steps}). Our experiments suggest that 1k steps of co-finetuning consistently lead to strong real-to-sim correlation, but even just a few hundred steps of finetuning can significantly improve correlation results. \Cref{fig:finetuning_steps} shows that too much finetuning can lead to overfitting and worsen correlation.

%% file: sections/06_discussion.tex
\section{Discussion}
In this work, we introduce \Acronym{}, a system for developing scalable evaluation benchmarks in simulation from real-world videos. \Acronym{} allows for personalized environment creation in simulation, corresponding to particular real-world environments. With a small amount of policy co-training on simulation data, these simulated environments can be made directly usable for policy evaluation. We perform a thorough empirical investigation, demonstrating strong correlation between simulated and real evaluations, out of the box even in unseen simulation environments. These simulation environments can serve as a useful tool for scalable evaluations of generalist policies. As advances are made in the field of 3D reconstruction and generation, the PolaRiS framework will benefit from higher fidelity environments and greater ease of use.

While a promising step for policy evaluation, there are still several points of improvement for \Acronym{}. Going beyond Gaussian splat rendering to richer diffusion based renderings can further bridge the real-to-sim visual gaps. Moreover, \Acronym{} performs a relatively simplistic identification of system dynamics. We can draw on a rich literature of system identification~\cite{systemID} to better identify the robot and environment dynamics. Lastly, we can consider hybrid simulation techniques, merging elements of world modeling with physics-based simulation to generate more scalable simulated evaluations, including more complex objects and scene configurations. 

%% file: sections/07_acks.tex
\section*{Acknowledgments}
This research was partially support by Amazon Science Hub, Army Research Lab, the DARPA TIAMAT Program, and the Toyota Research Institute University 2.0 Program, with compute support from Microsoft. The authors would like to thank Alex Fang, Ludwig Schmidt, Masha Itkina, Junjie Ye, Jiawei Yang for the early discussion on the project, Mateo Guaman Castro for help and feedback on figures and presentation, and Apurva Badithela and Lihan Zha for feedback on an earlier draft of this paper. We also thank Aditya Shah and Daniel Gorbatov for helping with evaluations, and Jenai (Xuning) Yang for advice and support with simulation.

%% file: sections/X_supp.tex
\setcounter{page}{1}

\appendix

\section{Additional Details for Preliminaries}

\subsection{2DGS}
\label{app:2dgs-prelim}
Given a set of calibrated RGB images \(\{I_i\}_{i=1}^{N}\) with known camera poses \(\{T_i\}_{i=1}^{N}\), 2DGS optimizes a set of primitives  
\[
\mathcal{D} = \bigl\{\,(\boldsymbol{\mu}_j,\; \mathbf{R}_j,\; \mathbf{S}_j,\; \mathbf{c}_j,\; \alpha_j)\bigr\}_{j=1}^{M},
\]
where \(\boldsymbol{\mu}_j \in \mathbb{R}^3\) is the disk centre, \(\mathbf{R}_j \in \mathrm{SO}(3)\) defines the disk’s orientation (via two tangent vectors) in world space, \(\mathbf{S}_j = \operatorname{diag}(s_{j,u},s_{j,v},0)\) encodes the elliptical variances in the disk plane, \(\mathbf{c}_j \in \mathbb{R}^3\) is the view-dependent colour and \(\alpha_j \in [0,1]\) the opacity. Rendering is carried out by computing a perspective-correct ray–splat intersection for each disk, projecting the elliptical footprint into image space, and then performing differentiable alpha-blending to produce a rendered image \(\hat I_i(\mathcal{D},T_i)\). 

The training objective is then given by a photometric loss plus geometry-regularisation terms:
\begin{align*}
    \mathcal{L}(\mathcal{D}) = &\sum_{i=1}^{N} \big\|\hat I_i(\mathcal{D},T_i) - I_i\big\|_2^2 \;\\
    &+\; \lambda_{\text{dist}}\,\mathcal{R}_{\text{dist}}(\mathcal{D}) \;+\; \lambda_{\text{norm}}\,\mathcal{R}_{\text{norm}}(\mathcal{D}).
\end{align*}

Here \(\mathcal{R}_{\text{dist}}\) is a \emph{depth} regularizer that encourages each disk to align closely along the viewing ray to reduce thickness and multi-view inconsistency, and \(\mathcal{R}_{\text{norm}}\) is a \emph{normal} regularizer that minimizes the discrepancy between disk normals and the gradient of the rendered depth map. Once optimized, the disks give rise to both high-quality novel-view synthesis and accurate geometry that can be extracted (e.g., via meshing) and used for simulation of robotic policies. While closely related to 3D Gaussian splatting \cite{3dgs}, the 2DGS algorithm more faithfully captures object geometries while maintaining high fidelity visuals. As we show in our experimental analysis, this ensures better zero-shot evaluation performance for robot policies trained in the real world. 

\section{Environment Generation with \Acronym{}}
\label{app:2dgs-method}

\subsection{Reconstruction}
To ensure all reconstructions are expressed in a consistent world coordinate frame, we define the ChArUco board origin as the canonical base frame $\mathcal{F}_0$. For each video, we get recovered camera poses $\{T_i\}_{i=1}^{N}$ for almost all video images in an arbitrary frame from COLMAP, and a subset of these frames will have measured camera coordinates $\{C_i\}_{i=1}^M; M\leq N$ with respect to the ChArUco board. Using the COLMAP's \emph{model aligner} functionality \cite{schonberger2016colmap}, we are able to transform the collective camera poses' arbitrary scale and orientation into an accurately scaled and oriented set of camera poses in the ChArUco board canonical base frame. 

\subsection{Scene Composition}
After training a Gaussian Splat and extracting a mesh, we turn all assets into USD format. For raytraced experiments, we baked the vertex colors from 2DGS mesh extraction into a texture. All scanned objects are also turned into USD format. Finally, scenes are put together using our scene composition GUI (Fig. \ref{fig:interface-viz}).

\section{Evaluation Tasks, Cotraining Hyperparameters, and Datasets}
\label{sec:app:tasks_and_hyperparameters}

\begin{figure}
    \centering
    \includegraphics[width=\linewidth]{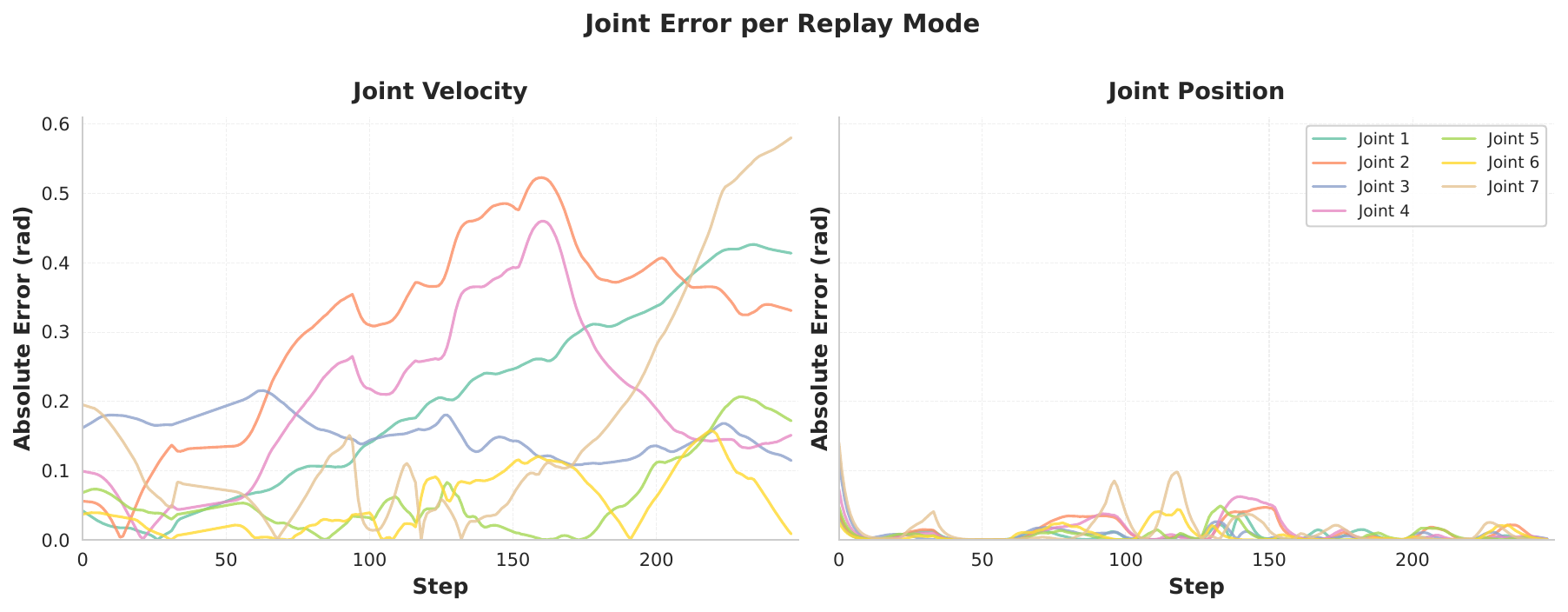}
    \caption{Comparison of absolute joint error overtime in open loop replay between joint velocity and joint position control modes.}
    \label{fig:openloopreplay}
\end{figure}

\begin{table*}[t]
\centering
\setlength{\tabcolsep}{6pt}        %
\renewcommand{\arraystretch}{1.35} %

\begin{tabular}{%
    >{\raggedright\arraybackslash}p{0.16\textwidth}%
    >{\raggedright\arraybackslash}p{0.30\textwidth}%
    >{\raggedright\arraybackslash}p{0.48\textwidth}}
\hline
\textbf{Task Name} & \textbf{Instruction} & \textbf{Step-by-Step Rubric} \\
\hline

Food Bussing &
Put all the foods in the bowl &
\begin{enumerate}[leftmargin=1.2em]
    \item Reach for any food item
    \item Lift the food item
    \item Place the first food item into the bowl
    \item Place the second food item into the bowl
\end{enumerate}
\\
\hline

Block Stacking &
Place and stack the blocks on top of the green tray &
\begin{enumerate}[leftmargin=1.2em]
    \item Reach for any block
    \item Lift the block
    \item Place any block on the tray
    \item Place the second block on the tray
    \item Second block is stacked on the first
\end{enumerate}
\\
\hline

Pan Cleaning &
Use the yellow sponge to scrub the blue handle frying pan &
\begin{enumerate}[leftmargin=1.2em]
    \item Reach for the yellow sponge
    \item Lift the sponge
    \item Move sponge into the pan
\end{enumerate}
\\
\hline

Move Latte Cup &
Put the latte art cup on top of the cutting board &
\begin{enumerate}[leftmargin=1.2em]
    \item Reach for the latte cup
    \item Pick up the cup
    \item Put it on top of the cutting board
\end{enumerate}
\\
\hline

Organize Tools &
Put the scissor into the large container &
\begin{enumerate}[leftmargin=1.2em]
    \item Reach for the scissor
    \item Pick up the scissor
    \item Put it into the container in large size
\end{enumerate}
\\
\hline

Tape Into Container &
Put the tape into the container &
\begin{enumerate}[leftmargin=1.2em]
    \item Reach for the tape
    \item Pick up the tape
    \item Put it into the container
\end{enumerate}
\\
\hline

\end{tabular}

\caption{Task prompts and the corresponding step-by-step rubric for the three evaluation tasks.
Scores are normalized between 0 and 1 for comparability across environments.}
\label{fig:task-prompts-rubric}
\end{table*}

\begin{table}[t]
\centering
\renewcommand{\arraystretch}{1.35}
\setlength{\tabcolsep}{14pt}

\begin{tabular}{l c}
\hline
\textbf{Hyperparameter} & \textbf{Parameter Value} \\
\hline
Action Space           & Joint Position (8-DoF robot) \\
Learning Rate Schedule & Cosine Decay \\
Warmup Steps           & 1{,}000 \\
Peak Learning Rate     & \(5 \times 10^{-5}\) \\
Decay Steps            & 1{,}000{,}000 \\
Decay Learning Rate    & \(5 \times 10^{-5}\) \\
Batch Size             & 128 \\
\hline
\end{tabular}
\caption{Shared hyperparameters used for all policies and experiments during co-training.}
\label{tab:shared-hyperparams}
\end{table}

\begin{table*}[t]
\centering
\renewcommand{\arraystretch}{1.3}

\begin{tabular}{%
    >{\raggedright\arraybackslash}p{0.22\textwidth}%
    >{\raggedright\arraybackslash}p{0.75\textwidth}%
}
\hline
\textbf{Model} & \textbf{Details} \\
\hline

pi05 & Action dim = 32, Action horizon = 15, no tokenizer \\

pi0\_fast & Action dim = 8, Action horizon = 10, FAST Tokenizer \\

pi0 & Action dim = 32, Action horizon = 10, no tokenizer \\

pi0\_100k & Action dim = 32, Action horizon = 10, no tokenizer \\

paligemma\_binning & Action dim = 8, Action horizon = 15, Binning Tokenizer \\
\hline
\end{tabular}

\caption{Policy-specific configurations for all DROID cotrained models. 
For LIBERO experiments, the only modification is changing action dimension to 7 for \textbf{pi0\_fast} and \textbf{paligemma\_binning}.}
\label{tab:policy_specific_hyperparams}
\end{table*}

\begin{table*}[t]
    \centering
    \setlength{\tabcolsep}{4pt}
    \renewcommand{\arraystretch}{1.3}

    \begin{tabularx}{\textwidth}{%
        >{\centering\arraybackslash}X
        >{\centering\arraybackslash}X
        >{\centering\arraybackslash}X}
    \hline
    \large \textbf{In-Domain} & \large \textbf{Near-Domain} & \large \textbf{OOD} \\
    \hline

    \includegraphics[width=\linewidth]{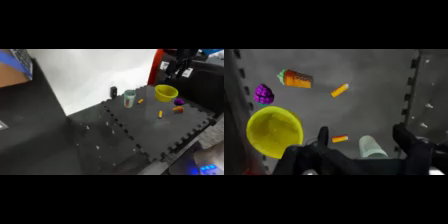} &
    \includegraphics[width=\linewidth]{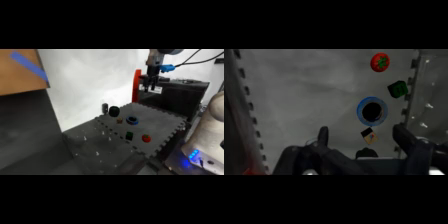} &
    \includegraphics[width=\linewidth]{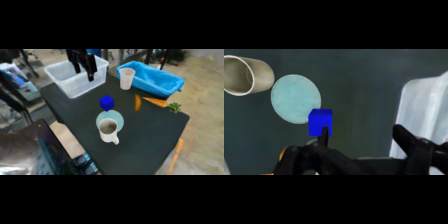}
    \\
    \hline

    Prompt: Put all the foods in the bowl &
    Prompt: Move the cubes from left to right &
    Prompt: Put the dishes into the plastic container
    \\
    \hline

    \begin{minipage}{\linewidth}\centering
        Environments: 18 \\
        Instructions: 21 \\
        Trajectories: 476 \\
        Timesteps: 79,077
    \end{minipage}
    &
    \begin{minipage}{\linewidth}\centering
        Environments: 18 \\
        Instructions: 21 \\
        Trajectories: 416 \\
        Timesteps: 65,489
    \end{minipage}
    &
    \begin{minipage}{\linewidth}\centering
        Environments: 15 \\
        Instructions: 18 \\
        Trajectories: 326 \\
        Timesteps: 56,041
    \end{minipage}
    \\
    \hline
    \end{tabularx}

    \caption{Details on the makeup of the different cotraining data mixes for ablating what kind of data is necessary to keep simulation evaluation faithful.}
    \label{table:data-mixes}
\end{table*}

\begin{table*}[t]
\centering
\setlength{\tabcolsep}{4pt}
\renewcommand{\arraystretch}{1.2}

\begin{tabular}{
    >{\centering\arraybackslash}p{0.25\textwidth} 
    >{\centering\arraybackslash}p{0.25\textwidth} 
    >{\centering\arraybackslash}p{0.25\textwidth} 
    >{\centering\arraybackslash}p{0.25\textwidth} 
}
\hline
\textbf{Splat} &
\textbf{Raytraced} &
\textbf{Untextured Environment} &
\textbf{Untextured Everything}
\\
\hline

\includegraphics[width=\linewidth]{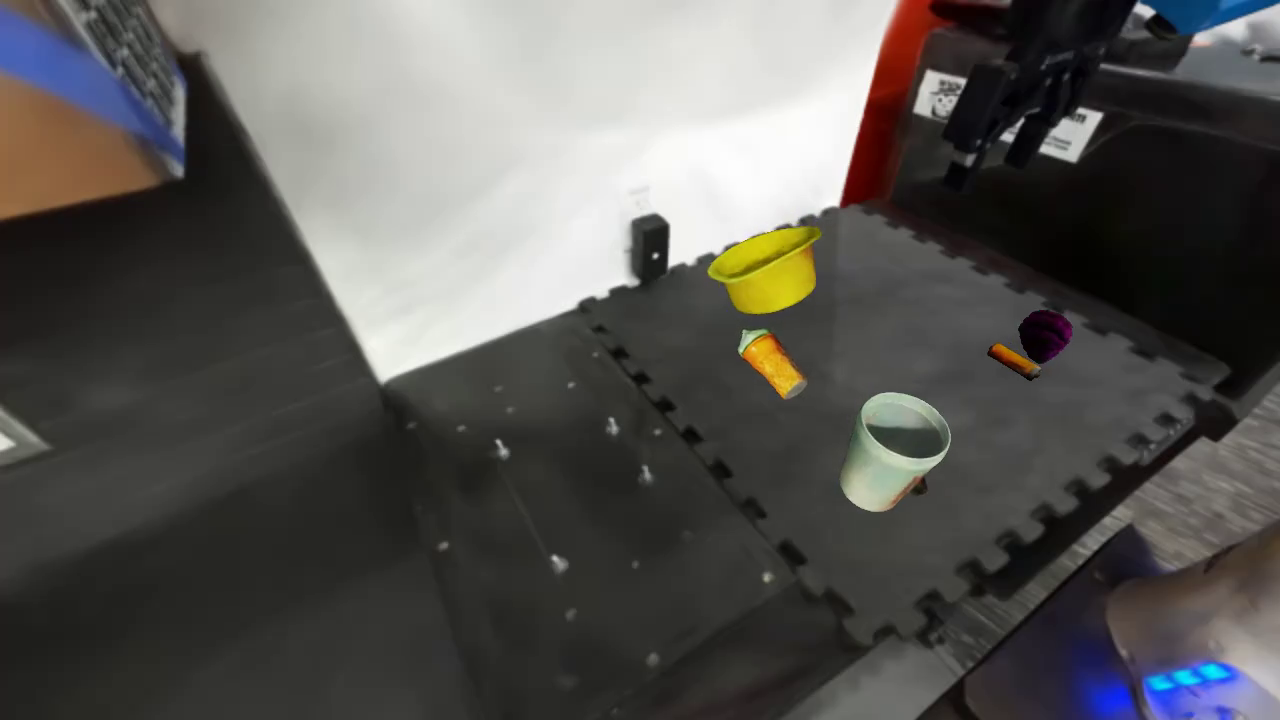} &
\includegraphics[width=\linewidth]{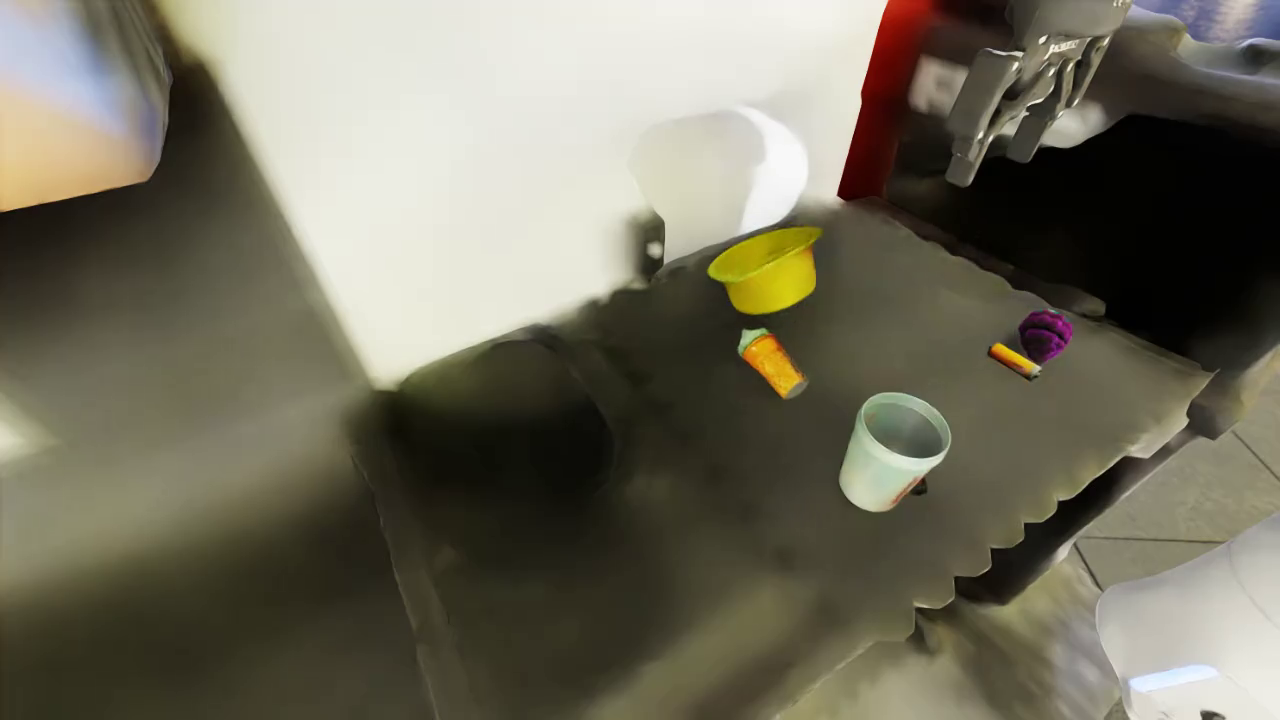} &
\includegraphics[width=\linewidth]{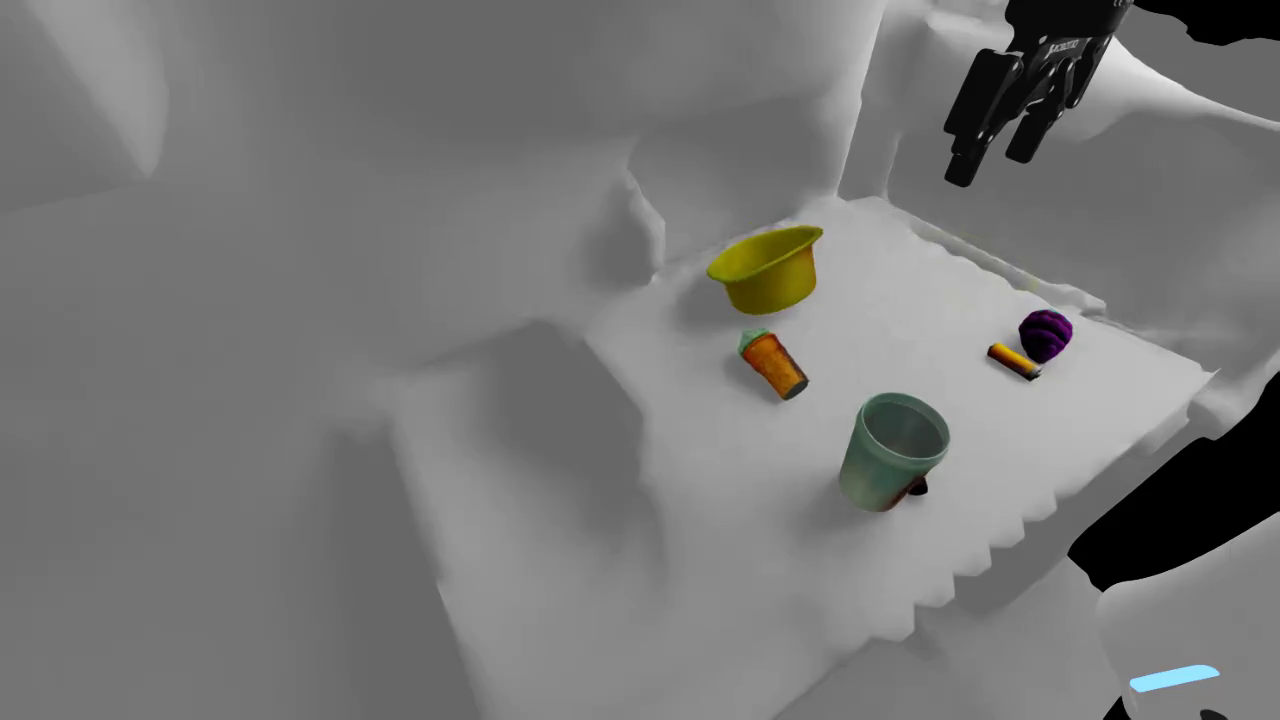} &
\includegraphics[width=\linewidth]{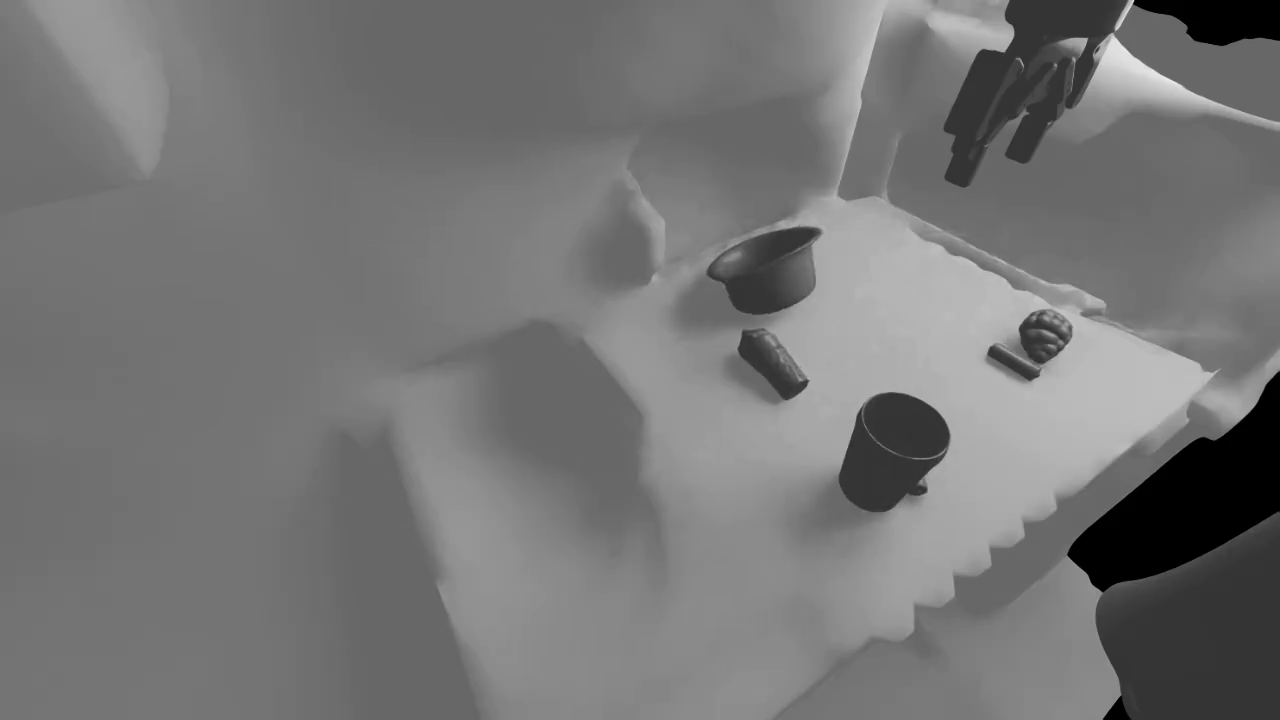}
\\
\hline
\end{tabular}

\caption{Examples of different visual fidelity renderings.}
\label{table:visual_fidelity}
\end{table*}

\subsection{Tasks}
We evaluate 6 tasks each in distinct environments.  Each task has its own criteria for partial progress scores so that we can evaluate policies with a more dense metric than sparse, binary success. All criteria are autonomously graded in simulation using privileged state information. For real world evaluations, a human watches each rollout and grades them according to the same rubric used in simulation. The criteria for each task in Table \ref{fig:task-prompts-rubric}. Each task may have a different score range, so scores and normalized between 0-1 to make allow for comparisons across environments.

\subsection{Cotraining}
For all of our experiments involving cotraining policies, we collected data in simulation environments using a twin of the real DROID platform \cite{khazatsky24droid}. To restrict behaviors from deviating significantly from the original DROID dataset, we adopted the same data collection scheme that was used in DROID. Data in simulation was collected with an Oculus Quest 2, and we setup our environments such that they could interface directly with the VR controller code shipped with the DROID code release. After data is collected, it's processed into the RLDS format, exactly matching the structure of the original DROID RLDS. Finally, policies were trained using the open-source DROID policy training code provided in \emph{openpi} \cite{openpi}, with a few modifications in dataloading. All cotraining was done with a mixture of \textbf{90\%} real DROID data and \textbf{10\%} simulated DROID data. Additional hyperparameters for cotraining can be found in Table \ref{tab:shared-hyperparams} and Table \ref{tab:policy_specific_hyperparams}. For our analysis of how many finetuning steps is needed to achieve best correlations, we ran independent finetuning runs with unique seeds (1-5). For each checkpoint we do 50 independent rollouts with the same initial conditions. All cotraining (both simulation and real world) is done on joint position action data. We find joint position policies to transfer much better from real to sim, with the intuition of joint position policies avoiding compounding error. We demonstrate this intuition in Fig.~\ref{fig:openloopreplay} using open loop replay comparing joint velocity and joint position policies.

\subsection{Datasets}
In our cotraining experiments, we performed some ablations on the differences in evaluation ability with different data mixtures, varying the closeness of the data to our test-time distribution. Specifically, our datasets were \textit{"OOD"}, \textit{"near-domain"}, and \textit{"in-domain"}. Details on examples and makeup of each dataset can be found in Table \ref{table:data-mixes}. We also performed ablations on how much visual fidelity mattered for faithful evaluation of policies. In this experiment, we had 4 datasets used for cotraining: \textit{splat}, \textit{raytraced}, \textit{untextured environment}, and \textit{untextured everything}. Examples of each can be seen in Table \ref{table:visual_fidelity}. All visual fidelity ablations were performed with \textit{OOD} data, with only differences in rendering. 

\section{Additional Experiments}
\label{app:additional-exps}

\subsection{Hardware-in-the-loop Experiments}
In order to determine what was causing the low correlation and performance in our real-to-sim environments, we set out to isolate and test the issues of visual gap and dynamics gap.
We accomplished this by running rollouts in the real world and selectively switching out parts of the rollout with simulation. For testing whether simulation dynamics was the issue, we fed the policy real observations and commanded joint positions with blocking control using the achieved simulation joint positions for every step (control frequency of 15 Hz for DROID). To test the visual gap, we matched up a scene between real and simulation and fed the policy simulation renders at each step. For control, we executed predicted actions on the real robot for $1/15$ seconds and commanded the achieved real joint positions to the simulation to create the effect of real dynamics with simulation observations -- evaluating success in simulation. All hardware-in-the-loop experiments were performed with the $\pi_0$ FAST DROID Joint Position policy. We found that introducing simulated dynamics with real visuals led to no drop in performance; however, introducing simulation visuals with real dynamics led to about a 25\% drop in performance measured across 3 tasks.

\begin{figure}
    \centering
    \includegraphics[width=0.8\linewidth]{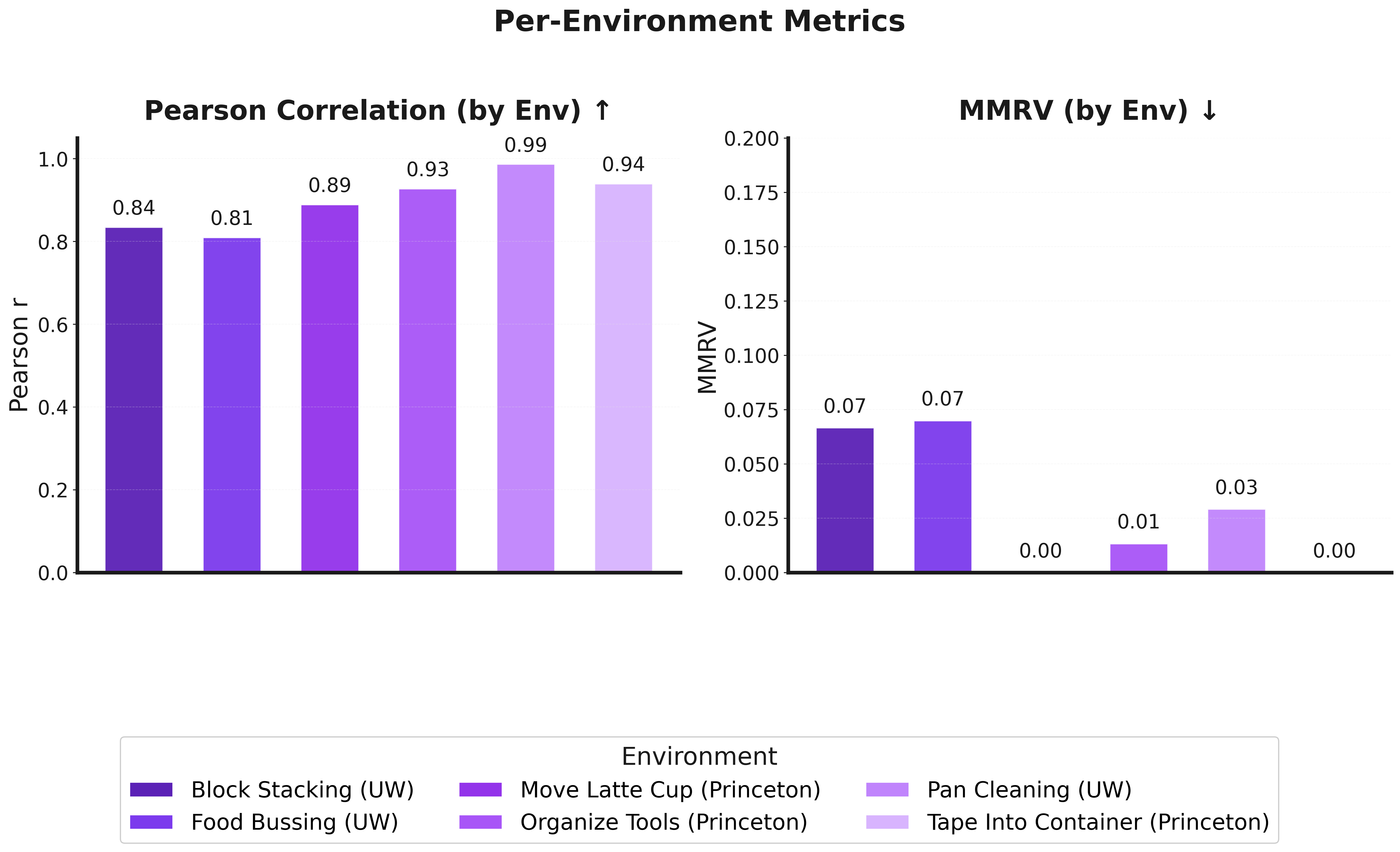}
    \caption{Correlation metrics for each individual environment, demonstrating PolaRiS's ability to faithfully evaluate policies in targeted environments for deployment in addition to general performance.}
    \label{fig:individual-correlation}
\end{figure}